\documentclass[a4paper]{cas-dc}

\usepackage[numbers]{natbib}

\usepackage{enumerate}
\usepackage{amsmath,amssymb,amsfonts}
\usepackage{mathtools}
\usepackage{mathrsfs}
\usepackage{algorithm}
\usepackage{algorithmic}
\usepackage{txfonts} 
\usepackage{longtable}
\usepackage{url}
\usepackage{graphicx}
\usepackage{subfigure} 
\usepackage{placeins}
\usepackage{caption}
\usepackage{xcolor} 
\usepackage{float}
\usepackage{subcaption}
\def\tsc#1{\csdef{#1}{\textsc{\lowercase{#1}}\xspace}}
\tsc{WGM}
\tsc{QE}
\tsc{EP}
\tsc{PMS}
\tsc{BEC}
\tsc{DE}

\begin{document}
\let\WriteBookmarks\relax
\def\floatpagepagefraction{1}
\def\textpagefraction{.001}

\title [mode = title]{\textcolor{black}{Short-Term Photovoltaic Forecasting Model for Qualifying Uncertainty during Hazy Weather}}
%

\author[1]{Xuan Yang}
\fnmark[1]{}
\credit{Conceptualization, Methodology, Data curation, Software, Writing - Original draft preparation}
\affiliation[1]{
    organization={School of Computer, Electronics and Information, Guangxi University},
    city={Nanning},
    postcode={530004}, 
    country={China}
}
\fntext[fn1]{These authors contributed to the work equally and should be regarded as co-first authors. }

\author[1]{Yunxuan Dong}[style=chinese]
\fnmark[1]
\credit{Data curation, Writing - Original draft preparation}


\author[1]{Lina Yang}[style=chinese]
\credit{Supervision, Writing – review \& editing}

\author[2]{Thomas Wu}[style=chinese]
\ead{wuxinzhangcs@outlook.com}

\affiliation[2]{
    organization={School of Electrical Engineering, Guangxi University},
    city={Nanning},
    postcode={530004}, 
    country={China}
}
\cormark[1]
\cortext[1]{Corresponding author at School of Electrical Engineering, Guangxi University, Nanning, China.}
\credit{Supervision, Writing – review \& editing}

\begin{abstract}
Solar energy is one of the most promising renewable energy resources.
Forecasting photovoltaic power generation is an important way to increase photovoltaic penetration.
\textcolor{black}{However, the difficulty in qualifying uncertainty of PV power generation, especially during hazy weather, makes forecasting challenging.
This paper proposes a novel model to address the issue. We introduce a modified entropy to qualify uncertainty during hazy weather while clustering and attention mechanisms are employed to reduce computational costs and enhance forecasting accuracy, respectively. Hyperparameters were adjusted using an optimization algorithm. Experiments on two datasets related to hazy weather demonstrate that our model significantly improves forecasting accuracy compared to existing models.}
\end{abstract}



\maketitle

\section{Introduction}

\begin{figure}[h]
    \centering
    \setlength{\abovecaptionskip}{0.cm}
    \includegraphics[width=1\linewidth]{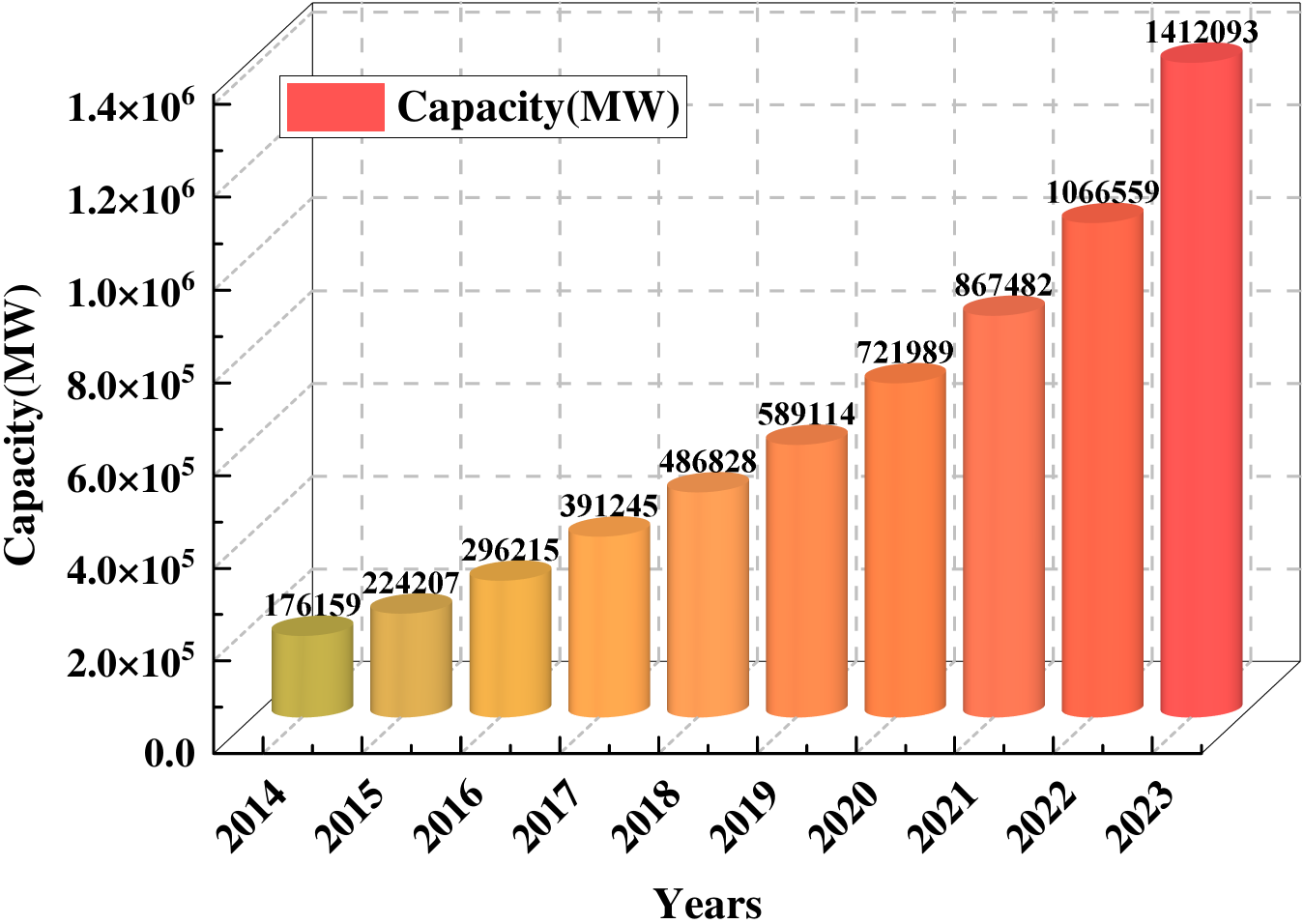}
    \caption{Global capacity of solar energy from 2014 to 2023.}
    \label{introduction}
    \vspace{-0.4cm} 
    \setlength{\abovecaptionskip}{-1cm} 
\end{figure}

\textcolor{black}{Fossil fuels extracted for decades to meet global energy demands, causing serious environmental problems \cite{song2024quantifying}. 
In contrast, solar power is known for its clean, pollution-free, and sustainable benefits \cite{yu2023short}.}
Fig. \ref{introduction} shows a significant increase in solar energy capacity over the past decade, with data derived from the International Renewable Energy Agency \cite{IRENA2024}.

Photovoltaic (PV) power generation embraces the property of uncertainty due to external factors \cite{zhang2024ultra}. 
\textcolor{black}{Consequently, this uncertainty decreases forecasting accuracy.
The power system may lead to serious failures if inaccurate PV forecasting is applied \cite{udagawa2017economic}.  
Consequently, accurate forecasting is essential to improve the stability of the power system \cite{gandhi2024value}.}

Existing PV forecasting models can be classified into physical, statistical, and hybrid models \cite{NGUYEN2022119603}. 
Physical models focus on theoretically explaining the works of PV power systems. \textcolor{black}{Statistical models use data analysis to make forecasts.}
Hybrid models combine multiple models to improve accuracy.
\textcolor{black}{Table \ref{classified model} lists existing PV forecasting models, showing their descriptions, advantages, and disadvantages.}

\textcolor{black}{PV panels receive less solar radiation because sunlight is scattered and particulate matter is deposited on the PV panels during hazy weather \cite{sadat2022review}. 
The maintenance of PV panels increases due to the deposition of particles on the panels.
The factors create uncertainty in PV power generation and affect power system stability \cite{zhang2020investigation}.
Therefore, it is essential to qualify uncertainty and make precise forecasts during hazy weather. 
Visibility drops below 10 kilometres due to light extinction by particulate matter in the atmosphere when the PM2.5 concentration reaches or exceeds 100 \(\text{$\mu$g/m}^3\). The weather can be considered as hazy weather under the circumstances \cite{guo2020temporal}.
In this study, we define weather with an air quality level of mild pollution, moderate pollution, or severe pollution as haze.}


\textcolor{black}{Existing works consider the impact of haze on PV forecasting \cite{ma2021new}.
Entropy is a way of qualifying uncertainty in time series \cite{jumarie2012derivation}. We create modified entropy to qualify uncertainty affected by haze. Then, we use the clustering method based on modified entropy to reduce computational costs. Lastly, we use attention mechanism to make forecasts.}
Drawing from the analysis above, the main contributions of this work are presented as follows:

\begin{itemize}
    \item \textcolor{black}{We create the Tsallis Entropy by Weighted Permutation Pattern (TEWPP) to qualify uncertainty during hazy weather.
    TEWPP examines the distribution of patterns to qualify uncertainty. This method reflects the distribution of PV power generation during hazy weather.}
    \item \textcolor{black}{We use hierarchical clustering based on entropy to reduce the computational costs of the model.
    This strategy can improve the similarity of elements in the same cluster, thereby reducing the computational costs.}
    \item \textcolor{black}{We use modified Retention Network (RetNet) to extract the features and make forecasts.}
    CNN uses kernel on input data to extract local features through multiplication and summation. 
    RetNet is used to make forecasts.
    The model improves the forecasting accuracy.
    \item We adopt the Non-dominated Sorting Genetic Algorithm-II (NSGA-II) to optimize the hyperparameters.
    NSGA-II searches for optimal solutions by simulating natural selection and genetic mechanisms.
    \textcolor{black}{The robustness of the model can be enhanced by using NSGA-II.}
    \item \textcolor{black}{We use datasets from a PV power station in a specific region of Jiangsu, China and another from Beijing, China.} The experimental results are evaluated through evaluation metrics, and these results are compared to others obtained from traditional models.
\end{itemize}

The remainder of this paper is organized as follows. 
Section \ref{related work} presents an overview of related works. 
Section \ref{Problem definition} introduces the optimal goal of the system model.
Section \ref{Proposed model} gives an architecture and a detailed description of CCRetNet. 
\textcolor{black}{Section \ref{experiments} provides an overview of the experimental setup, air quality analysis, and the results of single-step and multi-step forecasting experiments in Jiangsu Province and Beijing.}
Section \ref{conclusion} provides conclusions.

\begin{table*}[htbp]
\centering
\caption{Comparison of existing PV forecasting models}
\label{classified model}
\begin{tabular}{p{0.07\textwidth} p{0.3\textwidth} p{0.29\textwidth} p{0.24\textwidth}}
\toprule
Type & Description & Advantages & Disadvantages \\
\midrule
Physical models 
& Simulate the physical process of PV power plants to make forecasts.
& Provide high-accuracy forecasts. 
& Require large computational costs for measuring and updating data.
\\
\addlinespace
Statistical models 
& Build mathematical models with strict assumptions.
& Easy to explain the internal mechanisms. 
& Require input data to conform to the assumptions.\\
\addlinespace
Machine models 
& Make forecasts based on a huge amount of historical data. 
& Handle complex nonlinear relationships. 
& Difficult to explain the internal mechanisms. \\
\addlinespace
Hybrid models 
&Combine multiple models according to the characteristics of forecasting models to improve accuracy and reliability.
&Overcome limitations of a single model, improving accuracy and reliability.
& Result in higher computational costs than a single model. \\
\bottomrule
\end{tabular}
\end{table*}
\section{Related works} \label{related work}

PV forecasting is crucial for power systems \cite{yu2024tfeformer}.
Hence, scholar made most efforts in this regard \cite{zhou2023learning}, \cite{NGUYEN2022119603}, \cite{niu2024trend}, \cite{zhang2023incremental}. 
\textcolor{black}{In this section, we review related works from two aspects: weather factors and models.}

\textcolor{black}{Weather factors, particularly haze, are taken into account in the field of PV forecasting \cite{jia2022evaluation}.
Yao et al. \cite{yao2018support} introduced the air quality index and used the support vector machine algorithm to develop a new model for estimating solar radiation under haze conditions. The model improved accuracy and was suitable for regions with limited equipment or high pollution but did not explore using other neural network algorithms to assess the coupling effects of haze and weather parameters.
Fan et al. \cite{fan2018evaluating} employed a support vector machine model to improve global and diffuse solar radiation predictions by incorporating air quality index and various pollutants. The approach significantly enhanced accuracy, particularly in heavily polluted areas. However, as the number of input pollutants increases, the complexity and computational resource consumption also rise, with diminishing returns on improvement.
The research developed a new daily diffuse solar radiation model by incorporating the air quality index to improve accuracy in smog-affected areas \cite{yao2017research}. The air quality index adjustment significantly enhanced accuracy, especially in polluted regions like Beijing. However, the model's effectiveness varied across different regions, requiring further optimization for broader applicability.}

Nowadays, the attention mechanism is used in the field of PV forecasting \cite{qu2021day}.
The research used improved complementary empirical mode decomposition and variational mode decomposition for reducing noise \cite{yu2023short}. The attention mechanism focused on critical information to improve accuracy. The model was adjusted to data features to improve accuracy, whereas the complexity of the model limited its application.
\textcolor{black}{Yu et al. \cite{yu2024tfeformer} used temporal, frequency, and fourier attention to capture detailed time-series insights from PV power data.} The attention mechanism focused on crucial temporal and environmental features, thereby enhancing accuracy. However, this approach also obscured the significance of the time series itself.
Wang et al. \cite{wang2024efficient} combined a temporal convolutional network with an improved deep residual shrinkage network. The special attention mechanism in the model enhanced the ability to extract features in noisy environments. However, the model required significant computational resources.

\textcolor{black}{Table \ref{related works} summarizes other related works, including details of works, advantages, and disadvantages, where "Ref." denotes "References".}

\begin{table*}[t]
\centering
\caption{\textcolor{black}{List of related works}}\label{related works}
\begin{tabular}{p{0.017\textwidth} p{0.022\textwidth} p{0.32\textwidth} p{0.27\textwidth} p{0.25\textwidth}}
\toprule
\multicolumn{5}{l}{\textcolor{black}{PV forecasting  \textbf{considering weather factors}}} \\
\midrule
\textcolor{black}{Ref.} & \textcolor{black}{Year} & \textcolor{black}{Details} & \textcolor{black}{Advantages} & \textcolor{black}{Disadvantages} \\
\midrule

\textcolor{black}{\cite{ma2021new}}
&\textcolor{black}{2021}
&\textcolor{black}{Photovoltaic power; Power forecasting; Hazy 
conditions; Artificial neural network}
&\textcolor{black}{Enhanced accuracy in polluted environments.}
&\textcolor{black}{Relied on accuracy of air quality index data.}
\\

\textcolor{black}{\cite{fan2020hybrid}}
&\textcolor{black}{2020}
&\textcolor{black}{Air pollution; Prediction; Solar radiation}
&\textcolor{black}{Improved accuracy with the bat algorithm, enhancing convergence speed.}
&\textcolor{black}{Resulted in larger computation under different pollution levels.}
\\

\midrule

\multicolumn{5}{l}{\textcolor{black}{PV forecasting \textbf{compared model}}} \\
\midrule
\textcolor{black}{Ref.} & \textcolor{black}{Year} & \textcolor{black}{Details} & \textcolor{black}{Advantages} & \textcolor{black}{Disadvantages} \\
\midrule

\textcolor{black}{\cite{zhang2023improved}}
&\textcolor{black}{2023}
&\textcolor{black}{Photovoltaic power forecast; Channel attention mechanism; Time series forecasting}
&\textcolor{black}{Improved accuracy. Available for multivariate time series forecasting.}
&\textcolor{black}{Caused a risk of overfitting.} \\

\textcolor{black}{\cite{rai2023differential}}
&\textcolor{black}{2023}
&\textcolor{black}{Solar photovoltaic power; Short-term prediction; Deep learning; Attention net}
&\textcolor{black}{Enhanced processing capabilities through attention mechanisms.}
&\textcolor{black}{Sensitive to quality of input data.} \\

\textcolor{black}{\cite{huang2023memory}}
&\textcolor{black}{2023}
&\textcolor{black}{Temporal convolutional attention neural networks; PV power forecasting}
&\textcolor{black}{Captured the spatio-temporal features and improved the accuracy.}
&\textcolor{black}{Only represented the ultra-short-term prediction.}
\\

\bottomrule
\end{tabular}
\end{table*}

\begin{table*}[t]
\centering
\caption{List of abbreviations and parameters}
\begin{tabular}{p{0.085\textwidth} p{0.36\textwidth} p{0.085\textwidth} p{0.36\textwidth}}
\toprule
Abbreviations & Description & Abbreviations & Description \\
\midrule
MSR & Multi-Scale Retention &
PV & Photovoltaic\\
CCRetNet & Clustering and CNN-RetNet& 
Ref. & Reference \\
CNN & Convolutional Neural Network &  
RetNet & Retention Network \\
FLOPs & Floating Point Operations & 
RMSE & Root Mean Squared Error\\
FFN & Feed-Forward Network & 
RNN & Recurrent Neural Network\\
GRU & Gate Recurrent Unit & 
RS & Random Search\\
LSTM & Long Short Term Memory & 
TPE & Tree-structured Parzen Estimator\\
MAE & Mean Absolute Error &  
TEWPP & The Tsallis Entropy by Weighted Permutation Pattern   \\ 
\toprule  
Parameter & Description & Parameter & Description \\
\midrule

$k$ & The optimal number of PV power generation clusters & $\beta$ & The parameter for TEWPP \\
$\mathcal{L}_n(\cdot)$ & The loss function for the $n$-th cluster & $\boldsymbol{x}_{p}, \boldsymbol{x}_{q}$ & The series of PV power generation \\
$\phi_n$ & The parameters for the $n$-th model & $\boldsymbol{w}$ & The weight between the $\boldsymbol{x}_{p}$ and $\boldsymbol{x}_{q}$ \\
$\psi_n$ & The hyperparameters for the $n$-th model & $\boldsymbol{x}_{p}(n), \boldsymbol{x}_{q}(n)$ & The element of $\boldsymbol{x}_{p}$ and $\boldsymbol{x}_{q}$, respectively \\
$\Phi$ & The set of all model parameters & $d(\cdot,\cdot)$ & The distance \\
$\Psi$ & The set of all model hyperparameters & $d_n$ & Abbreviation for $d(\boldsymbol{x}_{p}(n),\boldsymbol{x}_{q}(n))$ \\
$\phi_n^*$ & The optimal parameters for the $n$-th model in training phase & $d_n'$ & Sorted $d_n$ \\
$\psi_n^*$ & The optimal hyperparameters for the $n$-th model in validating phase & $C_p$ & The $p$-th cluster \\
$\alpha$ & The weight parameter for training phase and validating phase & $d$ & Hidden dimension for RetNet \\
$\boldsymbol{x}$ & The series of PV power generation & $\boldsymbol{X^0},\boldsymbol{X}$ & The input data for RetNet \\
$T$ & The length of series for PV power generation $\boldsymbol{x}, \boldsymbol{x}_{p}$ and $ \boldsymbol{x}_{q}$ & $\boldsymbol{W_V}, \boldsymbol{W_Q},\boldsymbol{W_K}$ & The weight matrices \\
$\boldsymbol{x}^{m,\tau}_{t}$ & The segments of $\boldsymbol{x}$ & $\boldsymbol{V}, \boldsymbol{Q},\boldsymbol{K}$ & The projection \\
$m$ & The dimension of $\boldsymbol{x}^{m,\tau}_{i}$ & $\boldsymbol{v}_n, \boldsymbol{q}_n,\boldsymbol{k}_n$ & The projection \\
$\tau$ & The time lag of $\boldsymbol{x}^{m,\tau}_{i}$ & $\boldsymbol{s}_n$ & The state \\
$\pi^{m,\tau}_{j}$ & The unique permutation & $\boldsymbol{o}_n$ & Vectors $\boldsymbol{v}_n$ mapped through state $\boldsymbol{s}_n$ \\
$\omega_t$ & The weight parameter of $\boldsymbol{x}^{m,\tau}_{t}$ & $\gamma$ & Attenuation factor \\
$\mathbb{I}_A(\cdot)$ & The indicator function of set $A$ & $h$ & The number of retention heads in each layer \\
$\hat{\pi}(\boldsymbol{x})$ & The permutation of the series $\boldsymbol{x}$ & $d_h$ & The head dimension \\
\bottomrule
\end{tabular}
\end{table*}

\twocolumn 
\clearpage 

\begin{figure*}[htbp]
    \centering
    \includegraphics[width=0.94\linewidth]{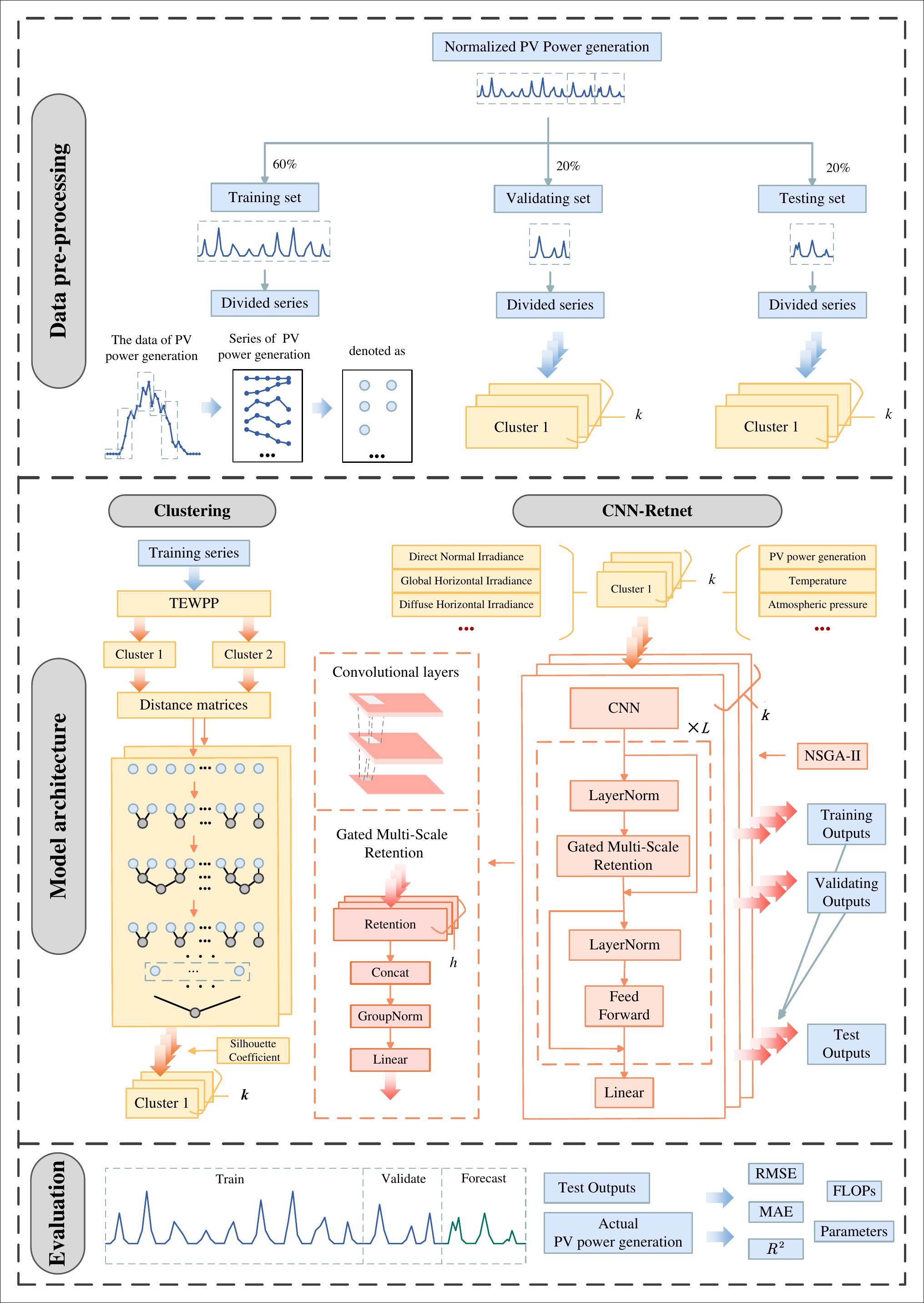}
    \caption{Framework of proposed model}
    \label{model architecture}
\end{figure*}
\clearpage 
\section{Problem definition} \label{Problem definition}
\textcolor{black}{Let $k$ represent the optimal number of clusters for PV power generation.
$\mathcal{L}_n(\cdot)$ denotes the loss function of the forecasting model for the $n$-th cluster.
The set of parameters is denoted by $\phi_n$, and the set of hyperparameters is denoted by $\psi_n$ for the $n$-th model.}
$\Phi=\{\phi_1,\phi_2,\cdots, \phi_k\}$ denotes the set of parameters for all model and $\Psi=\{\psi_1,\psi_2,\cdots, \psi_k\}$ denotes the set of hyperparameters for all model.

\textcolor{black}{During the training phase, the objective is to minimize the loss function for each cluster.} The objective function denotes as follows:
\begin{equation}
  \min_{\Phi} \sum_{n=1}^{k} \mathcal{L}_n(\phi_n).  \label{train}
\end{equation}

\textcolor{black}{The optimal set $\phi_n^{*}$ is determined during the training phase when equation (\ref{train}) obtains its minimum value} (i.e.,$\phi_n^{*}= \arg\min_{\Phi} \sum_{n=1}^{k} \mathcal{L}_n(\phi_n)$).

\textcolor{black}{For the validating phase, it should be assigned to the cluster that minimizes the loss function for the new data segment denoted by $\boldsymbol{x}$.} The objective function denotes as follows:
\begin{equation}
    \min_{n \in \{1, \ldots, k\}} \mathcal{L}_n(\boldsymbol{x},\psi_n; \phi_n^{*}) \label{validate}.
\end{equation}

\textcolor{black}{The optimal set $\psi_n^{*}$ is determined during the validating phase when equation (\ref{validate}) obtains its minimum value} (i.e., $\psi_n^{*}= \arg\min_{\Psi} \sum_{n=1}^{k} \mathcal{L}_n(\psi_n)$).

Combining both phases, the problem is defined as follows:
\begin{equation}
    \min \left( \alpha \sum_{n=1}^{k} \mathcal{L}_n(\phi_n) + (1 - \alpha) \sum_{n=1}^k \mathcal{L}_n(\boldsymbol{x},\psi_n; \phi_n^*) \right), \label{training and validating}
\end{equation}
\textcolor{black}{where $\alpha$ denotes a weighting parameter that balances minimizing the training loss and the validating loss.}

\textcolor{black}{Equation (\ref{training and validating}) ensures that the model of each cluster is well-trained (i.e., minimizing the loss function during training) and that new data is assigned to the most suitable cluster (i.e., minimizing the loss function during validating).}

\section{Proposed model} \label{Proposed model}
\subsection{Model architecture}

In this paper, we propose CCRetNet based on two crucial components: Clustering and CNN-RetNet. The architecture of the proposed model is demonstrated in Fig. \ref{model architecture}. \textcolor{black}{In brief, the aim of the proposed model is to qualify uncertainty, extract features, and improve forecasting accuracy during hazy weather. }

We create the Tsallis Entropy by Weighted Permutation Pattern (TEWPP) to qualify uncertainty.
\textcolor{black}{Based on TEWPP, hierarchical clustering is used to reduce the computational costs.} 
The hierarchical clustering utilizes the novel calculation of distance between series and the median linkage criterion as referenced in  \cite{He2022CATNCA}. 
\textcolor{black}{Additionally, we use modified RetNet to extract the features and make forecasts. CNN extracts the features, and RetNet is used to make forecasts.}
NSGA-II optimizes the hyperparameters to enhance the robustness.

\subsection{Clustering method}
\textcolor{black}{This section covers both TEWPP and hierarchical clustering.}
We first introduce TEWPP following the model architecture.
\subsubsection{The tsallis entropy by weighted
permutation pattern}

\textcolor{black}{The weighted permutation pattern is a method for time series analysis. Compared to permutation patterns, it better captures complex structures \cite{Weighted-permutation_entropy}.}

\textcolor{black}{Considering a PV power generation series $\boldsymbol{x}=\{\boldsymbol{x}(1),$ $\boldsymbol{x}(2),\ldots,\boldsymbol{x}(T)\}$ with length $T$, we obtain its segment $\boldsymbol{x}^{m,\tau}_{t} = \{\boldsymbol{x}(t), \boldsymbol{x}(t+\tau), \ldots, \boldsymbol{x}(t+(m-1)\tau)\}$ for $t = 1, 2, \cdots$, where $m$ and $\tau$ respectively represent the dimension and time lag.} 
Every $\boldsymbol{x}^{m,\tau}_{t}$ has $m!$ possible permutations, and it results in a unique permutation  $\pi^{m,\tau}_{j}$.

\textcolor{black}{Before defining the probability of the weighted permutation pattern $p_{\omega}(\pi^{m,\tau}{j})$, we first define $\bar{x}^{m,\tau}{t}$ and the weight value $\omega_t$.}

$\bar{x}^{m,\tau}_{t}$ is the arithmetic mean of $\boldsymbol{x}^{m,\tau}_{t}$.
For every $\boldsymbol{x}^{m,\tau}_{t}$, the weight value $\omega_t$ is defined as follows:
\begin{equation}
    \omega_{t} = \frac{1}{m}\sum_{k=1}^m \left[ \boldsymbol{x}(t+(k-1)\tau)  - \bar{x}^{m,\tau}_{t} \right] ^2.
\end{equation}

\textcolor{black}{The probability of the weighted permutation pattern is defined as follows \cite{Weighted-permutation_entropy.87.022911}:
\begin{equation}
   p_{\omega}(\pi^{m,\tau}_{j}) = \frac{\sum_{t \leq T-(m-1)} \mathbb{I}_{u:\hat{\pi}(u)=\pi_{j}} \left(\boldsymbol{x}^{m,\tau}_{t}\right) \omega_{t}}{\sum_{t \leq T-(m-1)} \omega_{t}},
\end{equation}
where $\mathbb{I}_A(\boldsymbol{x})$ denotes the indicator function of set $A$ defined as
$\mathbb{I}_A(\boldsymbol{x}) = 1$ if $\boldsymbol{x} \in A$ and $\mathbb{I}_A(\boldsymbol{x}) = 0$ if $\boldsymbol{x} \notin A$. $\hat{\pi}(\boldsymbol{x})$ is a permutation of the series $\boldsymbol{x}$.}

\textcolor{black}{We create TEWPP by applying the weighted permutation pattern to Tsallis entropy inspired by \cite{QIN2021111477entropy}.} The calculation of TEWPP is defined as follows:
\begin{align}
    &H_{\beta}(\boldsymbol{x}, m, \tau) = \frac{1}{\beta - 1} \left[ 1 - \sum_{j=1}^{m!} \left( p_{\omega}(\pi^{m,\tau}_{j}) \right)^{\beta} \right],  \notag \\ &\beta>0,\beta\neq 1, \label{Tsallis}
\end{align}
where $\beta$ is the parameter.
$H_{\beta}(\boldsymbol{x},m,\tau)$ reduce to Shannon entropy when $\beta \to 1$.
$H_{\beta}(\boldsymbol{x},m,\tau)$ is influenced by large probability events when the parameter was set to $\beta > 1$.
$H_{\beta}(\boldsymbol{x},m,\tau)$ is influenced by small probability events when the parameter was set to $\beta <1$.

\textcolor{black}{Based on TEWPP, we categorize the PV power series into clusters of large uncertainty and small uncertainty. Hierarchical clustering is used for the clusters of large uncertainty and small uncertainty, respectively.}
\subsubsection{Hierarchical clustering} \label{Hierarchical clustering}
The core of hierarchical clustering is to construct a structure based on the distance between series and the distance between clusters.
\textcolor{black}{Consequently, different distance measures can result in different clustering outcomes.
We will next present the calculations for the distance between series (equation (\ref{distance series})) and between clusters (equation (\ref{distance clusters})), followed by the hierarchical clustering process (algorithm \ref{cluster}) and the evaluation metric (equation (\ref{SC})).}

\textcolor{black}{Let $\boldsymbol{x}_{p}=\{\boldsymbol{x}_{p}(t)\}^T$ and $\boldsymbol{x}_{q}=\{\boldsymbol{x}_{q}(t)\}^T$ are series of PV power generation with T-dimension for $p\neq q$.} The weight $\boldsymbol{w} = \{w_n\}^T$ between the $\boldsymbol{x}_{p}$ and $\boldsymbol{x}_{q}$ is defined as below:
\begin{align}
    \boldsymbol{w}=\{w_n\mid &w_n = \max \{\frac{\boldsymbol{x}_p(n)}{(\sum_{m_1=1}^{T} \boldsymbol{x}_{p}(m_1)},\frac{\boldsymbol{x}_{q}(n)}{(\sum_{m_2=1}^{T} \boldsymbol{x}_q(m_2)}\}, \notag \\
    & n=1,2,\ldots,T\}.
\end{align}

The distance $d(\boldsymbol{x}_p(n),\boldsymbol{x}_q(n))$ between $\boldsymbol{x}_{p}(n)$ and $\boldsymbol{x}_{q}(n)$ is defined as follows:
\begin{equation}
    d(\boldsymbol{x}_{p}(n),\boldsymbol{x}_{q}(n))=w_n(\boldsymbol{x}_{p}(n)-\boldsymbol{x}_{q}(n))^2.
\end{equation}

For convenience, we write $d(\boldsymbol{x}_{p}(n),\boldsymbol{x}_{q}(n))$ as $d_n$. We sort the $\{d_n\}^T$ from largest to smallest to get the $\{d'_n\}^T$. Therefore $d'_n$ satisfies :
\begin{equation}
    d'_1\geq d'_2 \geq \cdots \geq d'_T.
\end{equation}

We define the calculation for the distance between series. 
\textcolor{black}{The calculation adopts the average of the $\{d'_n\}^T$ in the middle, thereby reducing the influence of extreme values.}
Hence, the distance between 
$\boldsymbol{x}_{p}$ and $\boldsymbol{x}_{q}$ is defined as follows:
\begin{align}
    d(\boldsymbol{x}_{p}, \boldsymbol{x}_{q}) = 
    \frac{1}{n_{Q_3}-n_{Q_2}+1}\sum_{n=n_{Q_2}}^{n_{Q_3}}d'_n,
    \label{distance series}
\end{align}
where the position index of the second quartile $Q_2$ and the third quartile $Q_3$ denote $n_{Q_2}$ and $n_{Q_3}$, respectively.

\textcolor{black}{Traditional distance calculations between clusters in hierarchical clustering include the single linkage criterion, complete linkage criterion, average linkage criterion, and median linkage criterion.} We use a novel linkage criterion as referenced in \cite{He2022CATNCA} to compute the distance between different clusters as follows:
\begin{equation}
     d_{\alpha_1,\alpha_2}(C_p, C_q) = 
    \frac{1}{|S_{\alpha_1(N),\alpha_2(N)}|}\sum_{(\boldsymbol{x}_{p}, \boldsymbol{x}_{q}) \in S_{\alpha_1(N),\alpha_2(N)}}d(\boldsymbol{x}_{p},\boldsymbol{x}_{q}), \label{distance clusters}
\end{equation}
where $S_{\alpha_1(N),\alpha_2(N)}$ is a subset of all possible pairs $(\boldsymbol{x}_{p},\boldsymbol{x}_{q}), \boldsymbol{x}_{p} \in C_p, \boldsymbol{x}_{q} \in C_q$. We define $S_{\alpha_1(N),\alpha_2(N)}$ as follows:
\begin{align}
    S_{\alpha_1(N),\alpha_2(N)}=\{(\boldsymbol{x}_{p},\boldsymbol{x}_{q}),\boldsymbol{x}_{p} \in C_p, \boldsymbol{x}_{q} \in C_q| \notag \\
    r_{(\alpha_1(N))} \leq d(\boldsymbol{x}_{p},\boldsymbol{x}_{q}) \leq r_{(\alpha_2(N))} \},
\end{align}
where $C_p$ and $C_q$ stand for different clusters. $N=|C_p||C_q|$, where $|C_p|$ means the number of elements in cluster $C_p$. $r_{(1)}\leq r_{(2)}\leq \ldots \leq r_{(N)}$
denotes a rank-ordered set of the distances $d(\boldsymbol{x}_{p},\boldsymbol{x}_{q})$. Obviously, $r_{(1)}$ and $r_{(N)}$ represent the minimum distance and the maximum distance, respectively.
\textcolor{black}{To more effectively group similar segments while minimizing the impact of extreme values of the segments, the parameter $\alpha_1(N)$ and $\alpha_2(N)$ are set to $N/2$ and $3N/4$} (i.e. $r_{\alpha_1(N)}$ and $r_{\alpha_2(N)}$ represent the second quartile $Q_2$ and the third quartile $Q_3$ of the distances $d(\boldsymbol{x}_{p},\boldsymbol{x}_{q})$ respectively).

We use hierarchical clustering according to the definitions of the above two distant calculations. \textcolor{black}{The process of hierarchical clustering is detailed in algorithm \ref{cluster}.}

\begin{algorithm}
\caption{Hierarchical clustering algorithm} \label{cluster}
\begin{algorithmic}[1]
\renewcommand{\algorithmicrequire}{\textbf{Input:}}
\renewcommand{\algorithmicensure}{\textbf{Output:}}
\REQUIRE Series $\mathcal{X}$, distance matrix $\mathcal{D}$ between series, linkage criterion
\ENSURE A set of clusters and a dendrogram representing the hierarchy of clusters
\STATE Initialize each series as a single cluster
\WHILE{number of clusters $> 1$}
    \STATE Calculate the distance between all pairs of clusters using the specified $\mathcal{D}$
    \STATE Merge the two closest clusters based on the linkage criterion
    \STATE Update the distance matrix based on the linkage criterion
\ENDWHILE
\STATE Construct the dendrogram based on the merge history
\STATE \textbf{return} Clusters and dendrogram
\end{algorithmic}
\end{algorithm}

We employ the silhouette coefficient \cite{rousseeuw1987silhouettes} to determine the most suitable number of clusters. \textcolor{black}{The silhouette coefficient is a way to evaluate the clustering effect by combining cohesion and separation.} The value spans from $-1$ to $1$. The closer silhouette coefficient of $\boldsymbol{x}_{p}$  is to 1, the more compact $\boldsymbol{x}_{p}$ is within its cluster.

Let $a(\boldsymbol{x}_p)$ denotes the cohesion of the cluster $C_p$ and
$b(\boldsymbol{x}_p)$ denotes the separation between clusters.
\textcolor{black}{$a(\boldsymbol{x}_p)$ is defined as mean distance from the segment $\boldsymbol{x}_p$ to all other segments in the same cluster.
$b(\boldsymbol{x}_p)$ denotes smallest mean distance from the segment $\boldsymbol{x}_p$ all segments in any other cluster.
The calculations of $a(\boldsymbol{x}_p)$ and $b(\boldsymbol{x}_p)$ are defined as follows:}
\begin{align}
    a(\boldsymbol{x}_p) = \frac{1}{|C_p|-1}\sum_{\boldsymbol{x}_{p_1} \in C_p,p\neq p_1}{d(\boldsymbol{x}_p,\boldsymbol{x}_{p_1})}, \notag \\
    b(\boldsymbol{x}_p) = \min_{C_p \neq C_q} \frac{1}{|C_q|} \sum_{\boldsymbol{x}_{q} \in C_q}{d(\boldsymbol{x}_p,\boldsymbol{x}_{q})}.
\end{align}

The silhouette coefficient $S(\boldsymbol{x}_p)$ for $\forall \boldsymbol{x}_p$ is calculated as follows:
\begin{align}
    S(\boldsymbol{x}_p) = \frac{b(\boldsymbol{x}_p) - a(\boldsymbol{x}_p)}{\max\{a(\boldsymbol{x}_p), b(\boldsymbol{x}_p)\}}.\label{SC}
\end{align}

The average of the silhouette coefficient for the entire dataset is calculated as follows:
\begin{align}
    S = \frac{1}{|\Omega_k|} \sum_{\boldsymbol{x}_p\in \Omega_k}S(\boldsymbol{x}_p) ,
\end{align}
where $\Omega_k$ represents the set of all the segments.

\subsection{CNN-RetNet}
\textcolor{black}{In this section, we use CNN-RetNet to make forecasts inspired by \cite{sun2023retentive}}.
\subsubsection{CNN}
\textcolor{black}{To extract features from PV power generation data, we add three convolutional layers before applying RetNet. The first convolutional layer maps the input data. The second layer extracts initial features, while the third layer captures higher-order features.}
\subsubsection{Retention}
Given input data $\boldsymbol{X}^0 \in \mathbb{R}^{l_x\times d}$ denoted $\boldsymbol{X}$ (i.e. $\boldsymbol{X}=\{\boldsymbol{x}_n\}^{l_x}$ and $\boldsymbol{x}_n \in \mathbb{R}^{1\times d}$), where $d$ denotes hidden dimension. Let the weight define  
$\boldsymbol{W_V}\in \mathbb{R}^{d\times d}$. We project it to a one-dimensional function as follows:
\begin{equation}
    \boldsymbol{v}_n=\boldsymbol{x}_n \cdot \boldsymbol{W_V},
\end{equation}
where vector $\boldsymbol{v}_n \in \mathbb{R}^{1\times d}$.
$\boldsymbol{W_Q},\boldsymbol{W_K}\in \mathbb{R}^{d \times d}$ denote the weights that need to be learned. Similarly, the projections of $\boldsymbol{Q}$, $\boldsymbol{K}$, $\boldsymbol{q}_n$ and $\boldsymbol{k}_n$ are defined as follows:
\begin{align}         &\boldsymbol{Q}=\boldsymbol{XW_Q},\boldsymbol{K}=\boldsymbol{XW_K}, \\
&\boldsymbol{q}_n = \boldsymbol{x}_n\cdot\boldsymbol{W_Q}\in \mathbb{R}^{1 \times d}, \boldsymbol{k}_n = \boldsymbol{x}_n\cdot\boldsymbol{W_K} \in \mathbb{R}^{1 \times d}.
\label{QK}
\end{align}

Consider $\boldsymbol{o}_n$ is mapped to a vector $\boldsymbol{v}_n$ by state $\boldsymbol{s}_n \in \mathbb{R}^{d\times d}$.
The state $\boldsymbol{s}_n$ and the vector $\boldsymbol{o}_n$ are defined as follows:
\begin{align}
  & \boldsymbol{s}_n = \boldsymbol{B} \boldsymbol{s}_{n-1} + \boldsymbol{k}^T_n \boldsymbol{v}_n, \notag \\
  & \boldsymbol{o}_n = \boldsymbol{q}_n\boldsymbol{s}_n = \sum_{s=1}^n \boldsymbol{q}_n \boldsymbol{B}^{n-s} \boldsymbol{k}^T_s \boldsymbol{v}_s,
  \label{sn,on}
\end{align}
\textcolor{black}{where $\boldsymbol{B}\in \mathbb{R}^{d \times d}$ denotes a diagonal matrix, defined as $\boldsymbol{B}=\boldsymbol{\Lambda}(\gamma e^{i\theta})\boldsymbol{\Lambda}^{-1}$.} With the equation (\ref{QK}), equation (\ref{sn,on}) and the definition of $\boldsymbol{\Lambda}$, we can rewrite the expression of $\boldsymbol{o}_n$ as follows:
\begin{align}
    \boldsymbol{o}_n &= \sum_{s=1}^{n} \boldsymbol{q}_n (\gamma e^{i\theta})^{n-s} \boldsymbol{k}_s^T\boldsymbol{v}_s \notag \\
    &= \sum_{s=1}^{n}(\boldsymbol{q}_n (\gamma e^{i\theta})^{n}) (\boldsymbol{k}_s(\gamma e^{i\theta})^{-s})^T\boldsymbol{v}_s,
    \label{on}
\end{align}
where $\boldsymbol{q}_n (\gamma e^{i\theta})^{n}$, $\boldsymbol{k}_s(\gamma e^{i\theta})^{-s}$ are relative position embeddings proposed by Transformer \cite{islam2023comprehensive}.
\textcolor{black}{Additionally, when $\gamma$ is a scalar, equation (\ref{on}) can be further simplified as follows:}
\begin{equation}
    \boldsymbol{o}_n = \sum_{s=1}^{n} \gamma^{n-s} (\boldsymbol{q}_n e^{in\theta})(\boldsymbol{k}_s e^{-is\theta})^{\dagger}\boldsymbol{v}_s,
\end{equation}
where $\dagger$ is the conjugate transpose.

\begin{figure*}
    \centering
    \includegraphics[width=0.82\textwidth]{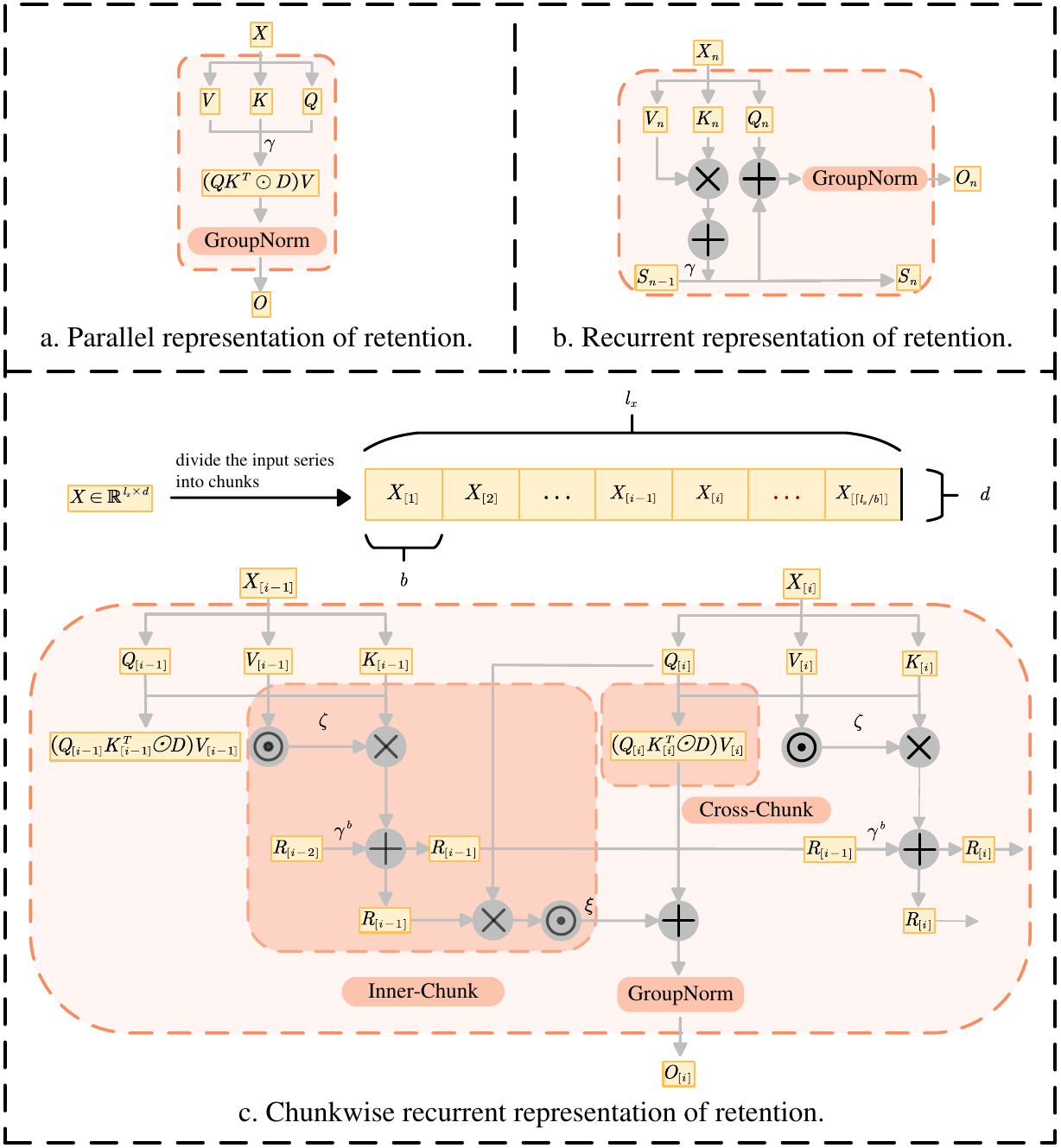}
    \caption{Three forms of retention}
    \label{retention}
\end{figure*}
\paragraph{The parallel representation}
The equation (\ref{parallel}) represents the parallel of retention.
\begin{equation}
\boldsymbol{Q} = (\boldsymbol{XW_Q}) \odot \Theta, \boldsymbol{K} = (\boldsymbol{XW_K}) \odot \overline{\Theta}, \boldsymbol{V} = \boldsymbol{XW_V}, 
\label{QKV}
\end{equation}
\begin{align}
&\Theta_n = e^{in\theta}, \quad D_{ns} = 
\begin{cases}
\gamma^{n-s}, & n \geq s \\
0, & n < s
\end{cases} \notag \\
&\text{Retention}(\boldsymbol{X},\gamma) = \boldsymbol{(QK^T \odot D)V},
\label{parallel}
\end{align}
\textcolor{black}{where $\odot$ denotes the two matrices are multiplied element by element.} $\Theta=\{\theta_n\}^{l_x}$ and $\overline{\Theta}$ denotes the complex conjugate of $\Theta$.
\paragraph{The recurrent representation of retention}
The equation (\ref{recurrent}) represents the recurrent of retention.
\begin{align}
    &\boldsymbol{S}_n = \gamma \boldsymbol{S}_{n-1} + \boldsymbol{k}_n^T \boldsymbol{v}_n, \notag \\
    &\text{Retention}(\boldsymbol{x}_n,\gamma) = \boldsymbol{q}_n\boldsymbol{S}_n, \quad n = 1, \ldots, l_x,
    \label{recurrent}
\end{align}
where the definition of $\boldsymbol{S}_n$ is similar to the state $\boldsymbol{s_n}$. $\boldsymbol{q}_n$, $\boldsymbol{k}_n$ and $\boldsymbol{v}_n$ can be derived from equation (\ref{QKV}).
\paragraph{The chunkwise recurrent representation of retention }

\textcolor{black}{We divide the input into chunks and apply the parallel representation specified in equation (\ref{parallel}) for computations within each chunk. In contrast, cross-chunk information is handled using the recurrent representation described in equation (\ref{recurrent}). Here, $b$ represents the chunk length. The retention output for the $i$-th chunk is calculated as follows:}

\begin{align}
&\boldsymbol{Q}_{[i]} = \boldsymbol{Q}_{bi:b(i+1)},  \boldsymbol{K}_{[i]} = \boldsymbol{K}_{bi:b(i+1)}, \boldsymbol{V}_{[i]} = \boldsymbol{V}_{bi:b(i+1)}, \notag\\
&\boldsymbol{R}_i = \boldsymbol{K}^T_{[i]}(\boldsymbol{V}_{[i]} \odot \zeta) + \gamma^b \boldsymbol{R}_{i-1}, \zeta_{i} = \gamma^{b-i-1}, \notag\\
&\text{Retention}(\boldsymbol{X}_{[i]},\gamma) = \underbrace{(\boldsymbol{Q}_{[i]} \boldsymbol{K}^T_{[i]} \odot \boldsymbol{D}) \boldsymbol{V}_{[i]}}_{\text{Inner-Chunk}} + \underbrace{(\boldsymbol{Q}_{[i]} \boldsymbol{R}_{i-1}) \odot \xi}_{\text{Cross-Chunk}},\notag\\
&\xi_{i} = \gamma^{i+1},
\end{align}
where the definition of $\boldsymbol{R}_i$ is similar to the state $\boldsymbol{s}_n$.
\textcolor{black}{The three forms of retention correspond to a, b, and c in figure \ref{retention} respectively.}

\subsubsection{Gated multi-scale retention}
$h$ retention heads are employed in each layer, and $d_h$ satisfies $h=d/ d_h$, where $d_h$ denotes head dimension. Each heads operates on distinct paramater matrices $\boldsymbol{W_Q}, \boldsymbol{W_K}, \boldsymbol{W_V} \in \mathbb{R}^{d\times d}$ and $\gamma$.
\textcolor{black}{The output of the MSR layer for the input $\boldsymbol{X}$ is defined as follows:}
\begin{align}
    \gamma &= 1 - 2^{-5\text{--}arange(0,h)} \in \mathbb{R}^{h}, \notag\\
    \text{head}_i &= \text{Retention}(\boldsymbol{X}, \gamma_i), \notag \\
    \boldsymbol{Y} &= \text{GroupNorm}_h(\text{Concat}(\text{head}_1, \ldots, \text{head}_{h})), \notag \\
    \text{MSR}(\boldsymbol{X}) &= (\text{swish}(\boldsymbol{XW_G}) \odot \boldsymbol{Y})\boldsymbol{W_O}, \label{MSR}
\end{align}
where $\boldsymbol{W_G, W_O}\in \mathbb{R}^{d\times d}$ are learnable parameters. Swish is the activation function. Since there is a different $\gamma$ for each head, the output of each head needs to be normalized separately and then combined. 

\subsubsection{Retention network}
In an $L$-layer Retention Network, we integrate multi-scale retention (MSR) with the feed-forward network (FFN). Let $W_1$ and $W_2$ be parameters. FFN is calculated as follows:
\begin{equation}
  \text{FFN}(\boldsymbol{X})=\text{gelu}(\boldsymbol{X}W_1)W_2, \label{FFN}
\end{equation}
\textcolor{black}{where gelu denotes the activation function.}

The calculation for the $l$-th layer of RetNet is denoted as follows: 
\begin{align}
    \boldsymbol{Y}^l&=\text{MSR}(\text{LayerNorm}(\boldsymbol{X}^l))+\boldsymbol{X}^l\\
    \boldsymbol{X}^{l+1}&=\text{FFN}(\text{LayerNorm}(\boldsymbol{Y}^l))+\boldsymbol{Y}^l
\end{align}
where $l \in [ 0, L ]$. The output denotes $\boldsymbol{X}^L$. The process of RetNet is shown in algorithm \ref{RetNet}.

\begin{algorithm}
\caption{RetNet Algorithm}
\begin{algorithmic}[1] \label{RetNet}
\renewcommand{\algorithmicrequire}{\textbf{Input:}}
\renewcommand{\algorithmicensure}{\textbf{Output:}}
\REQUIRE The input data $\boldsymbol{X}$.
\ENSURE The PV power generation data.
\FOR{each layer $l = 1$ to $L$}
    \FOR{each head $i = 1$ to $h$}
        \STATE $\text{head}_i \leftarrow \text{Retention}(\boldsymbol{X}, \gamma_i)$
    \ENDFOR
    \STATE $\boldsymbol{Y} \leftarrow \text{GroupNorm}_h(\text{Concat}(\text{head}_1, \ldots, \text{head}_h))$
    \STATE $\text{MSR}(\boldsymbol{X}) \leftarrow (\text{swish}(\boldsymbol{XW_G}) \odot \boldsymbol{Y})\boldsymbol{W_O}$
    \STATE $\boldsymbol{Y}^l \leftarrow \text{MSR}(\text{LayerNorm}(\boldsymbol{X}^l)) + \boldsymbol{X}^l$
    \STATE $\boldsymbol{X}^l \leftarrow \text{FFN}(\text{LayerNorm}(\boldsymbol{Y}^l)) + \boldsymbol{Y}^l$
\ENDFOR
    \STATE $\text{Output} \leftarrow X_L$
\end{algorithmic}
\end{algorithm}

\subsection{Optimization}
\textcolor{black}{NSGA-II is used in multi-objective optimization and is known for finding a balanced set of solutions. It is commonly applied in engineering, economics, logistics, and other fields.}  In this study, we utilize the NSGA-II algorithm.
\section{Experiments} \label{experiments}

\subsection{Experimental setup}

\subsubsection{Datasets}

\textcolor{black}{Both datasets are multi-sample, multivariate, and exhibit temporal variation. 
One dataset, which contains eight variables, originates from a 2015 PV power station in a specific region of Jiangsu Province, China. 
The other dataset originates from Beijing, China, collected in 2019 and contains nine variables. 
Table \ref{table:jiangsuvariables} details the variables in the 2015 Jiangsu dataset, while Table \ref{table:beijingvariables} outlines the variables in the 2019 Beijing dataset.}


\subsubsection{Implementation details}
\textcolor{black}{We apply min-max normalization to each variable at different scales.}
The model generates outputs from normalized input data, and outputs are inversely normalized for evaluation.

In calculating TEWPP, we set $ m=5$, $ \tau=2$, and $ \beta = 0.8$. We employ the silhouette coefficient to determine the optimal number of clusters. \textcolor{black}{The detailed introduction of the silhouette coefficient has already been explained in Section \ref{Hierarchical clustering}.} The hyperparameters are adjusted using the Optuna optimal framework.

\begin{table*}[!H]
\centering
\caption{\textcolor{black}{Statistical table of PV power generation and meteorological variables in a specific region of Jiangsu Province, 2015}}
\begin{tabular}{p{0.3\textwidth} p{0.1\textwidth} p{0.1\textwidth} p{0.1\textwidth} p{0.1\textwidth}}
\toprule
\textcolor{black}{\textbf{Variable Name}} &\textcolor{black}{\textbf{Unit}} & \textcolor{black}{\textbf{Max Value}} & \textcolor{black}{\textbf{Min Value}} & \textcolor{black}{\textbf{Mean Value}} \\
\midrule
\textcolor{black}{Direct Normal Irradiance} & \textcolor{black}{W/m$^2$} & \textcolor{black}{1051.68} & \textcolor{black}{0.00} & \textcolor{black}{210.40} \\
\textcolor{black}{Global Horizontal Irradiance} & \textcolor{black}{W/m$^2$} & \textcolor{black}{962.29} & \textcolor{black}{0.00} & \textcolor{black}{192.52} \\
\textcolor{black}{Diffuse Horizontal Irradiance} & \textcolor{black}{W/m$^2$} & \textcolor{black}{94.65} & \textcolor{black}{0.00} & \textcolor{black}{18.94} \\
\textcolor{black}{Temperature of Component} & \textcolor{black}{$^{\circ}$C} & \textcolor{black}{53.00} & \textcolor{black}{-6.30} & \textcolor{black}{16.65} \\
\textcolor{black}{Ambient Temperature} & \textcolor{black}{$^{\circ}$C} & \textcolor{black}{36.50} & \textcolor{black}{-4.70} & \textcolor{black}{15.44} \\
\textcolor{black}{Atmospheric Pressure} & \textcolor{black}{hPa} & \textcolor{black}{1040.40} & \textcolor{black}{994.00} & \textcolor{black}{1017.52} \\
\textcolor{black}{Relative Humidity} & \textcolor{black}{\%} & \textcolor{black}{100.00} & \textcolor{black}{15.00} & \textcolor{black}{76.82} \\
\textcolor{black}{PV Power Generation} & \textcolor{black}{MW} & \textcolor{black}{115.775} & \textcolor{black}{-0.44} & \textcolor{black}{17.68} \\
\midrule
\end{tabular}
\label{table:jiangsuvariables}
\end{table*}
\begin{table*}[!h]
\centering
\caption{\textcolor{black}{Statistical table of PV power generation and meteorological variables in Beijing, 2019}}
\begin{tabular}{p{0.3\textwidth} p{0.1\textwidth} p{0.1\textwidth} p{0.1\textwidth} p{0.1\textwidth}}
\toprule
\textcolor{black}{\textbf{Variable Name}} & \textcolor{black}{\textbf{Unit}} & \textcolor{black}{\textbf{Max Value}} & \textcolor{black}{\textbf{Min Value}} & \textcolor{black}{\textbf{Mean Value}} \\
\midrule
\textcolor{black}{Air Temperature} & \textcolor{black}{$^{\circ}$C} & \textcolor{black}{40.973}   & \textcolor{black}{-16.618} & \textcolor{black}{12.922} \\
\textcolor{black}{Precipitation} & \textcolor{black}{mm/hour} & \textcolor{black}{7.493} & \textcolor{black}{0.000}   & \textcolor{black}{0.056}  \\
\textcolor{black}{Snowfall} & \textcolor{black}{mm/hour} & \textcolor{black}{3.242}    & \textcolor{black}{0.000}   & \textcolor{black}{0.002} \\
\textcolor{black}{Snow Mass} & \textcolor{black}{kg/m$^2$} & \textcolor{black}{7.001}    & \textcolor{black}{0.000}   & \textcolor{black}{0.148} \\
\textcolor{black}{Air Density} & \textcolor{black}{kg/m$^3$} & \textcolor{black}{1.378}    & \textcolor{black}{1.093}   & \textcolor{black}{1.212}  \\
\textcolor{black}{Ground-level Solar Irradiance} & \textcolor{black}{W/m$^2$} & \textcolor{black}{1012.688} & \textcolor{black}{0.000}   & \textcolor{black}{191.877} \\
\textcolor{black}{Top of Atmosphere Solar Irradiance} & \textcolor{black}{W/m$^2$} & \textcolor{black}{1259.022} & \textcolor{black}{0.000}   & \textcolor{black}{328.785} \\
\textcolor{black}{Cloud Cover Fraction} & \textcolor{black}{[0, 1]} & \textcolor{black}{0.999}    & \textcolor{black}{0.000}   & \textcolor{black}{0.351}  \\
\textcolor{black}{PV Power Generation} & \textcolor{black}{MW} & \textcolor{black}{0.875}    & \textcolor{black}{0.000}   & \textcolor{black}{0.177} \\
\bottomrule
\end{tabular}
\label{table:beijingvariables}
\end{table*}

\subsubsection{Evaluation metrics}
\textcolor{black}{The objective of PV forecasting is to improve accuracy while reducing computational costs. Consequently, we use three evaluation metrics for accuracy and two for computational costs.}

\textcolor{black}{There are three evaluation metrics for accuracy: the Root Mean Squared Error (RMSE), the Mean Absolute Error (MAE), and the coefficient of determination $R^2$.} Each of them offers a unique perspective. RMSE focuses on larger values, MAE focuses on median error, and Coefficient of determination $R^2$ focuses on the proportion of variance in the dependent variable forecasting from the independent variables. The calculations of metrics are as follows:

\begin{equation}
    \text{RMSE} =\left( \frac{1}{M} \sum_{i=1}^{M} (y_i - \hat{y}_i)^2 \right)^{\frac{1}{2}},
\end{equation}
\begin{equation}
    \text{MAE} = \frac{1}{M} \sum_{i=1}^{l} |y_i - \hat{y}_i|,
\end{equation}
\begin{equation}
    R^2 = 1 - \frac{\sum_{i=1}^{M}(y_i - \hat{y}_i)^2}{\sum_{i=1}^{M}(y_i - \bar{y})^2},
\end{equation}
where $y_i$ denotes the actual value, $\hat{y}_i$ denotes the forecasting value. $\bar{y}$ denotes the mean of the actual values. $M$ denotes the number of samples. 

\textcolor{black}{There are two evaluation metrics for computational costs: floating-point operations (FLOPs) and the number of parameters. FLOPs measure the number of floating-point operations required by the model during the inference process, while the number of parameters evaluates the model's size and storage requirements.}

\subsubsection{Computer environment} 
The experiments are conducted on two distinct setups.  One includes a GeForce RTX 4090 GPU and a 13th Gen Intel(R) Core(TM) i9-13900KF processor on Ubuntu 22.04.3 LTS. Another includes a GeForce RTX 4060Ti GPU and a 13th Gen Intel(R) Core(TM) i5-13400F processor on Windows 11.
\textcolor{black}{Both configurations use Python 3.10 and the Pytorch deep learning framework.}

\begin{figure*}[H]
    \centering
    \subfigure[\textcolor{black}{Monthly distribution of air quality days in a specific region of Jiangsu Province, 2015}]{
        \includegraphics[width=0.47\textwidth]{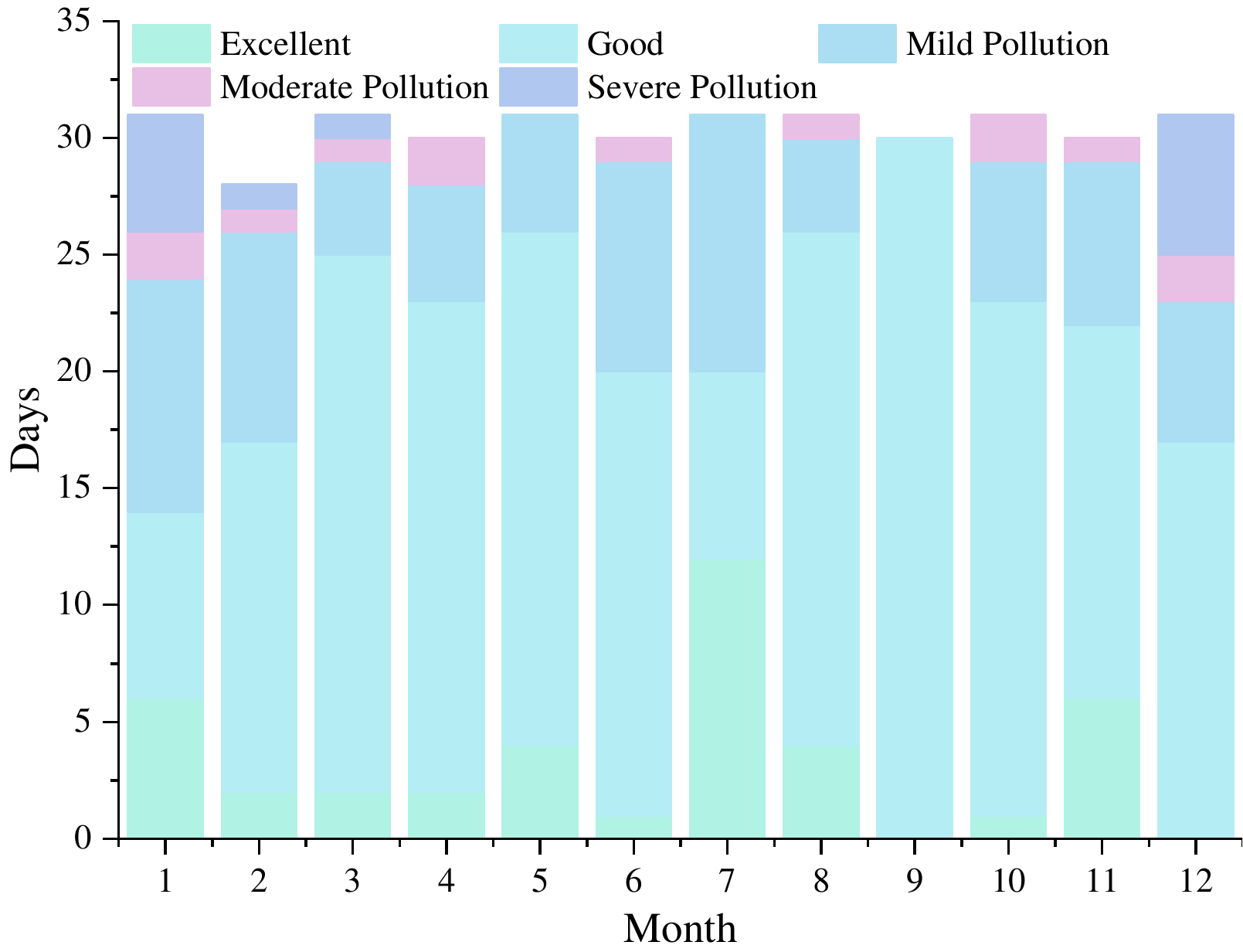}
        \label{AQIyan}
    }
    \subfigure[\textcolor{black}{Monthly distribution of air quality days in Beijing, 2019}]{
        \includegraphics[width=0.47\textwidth]{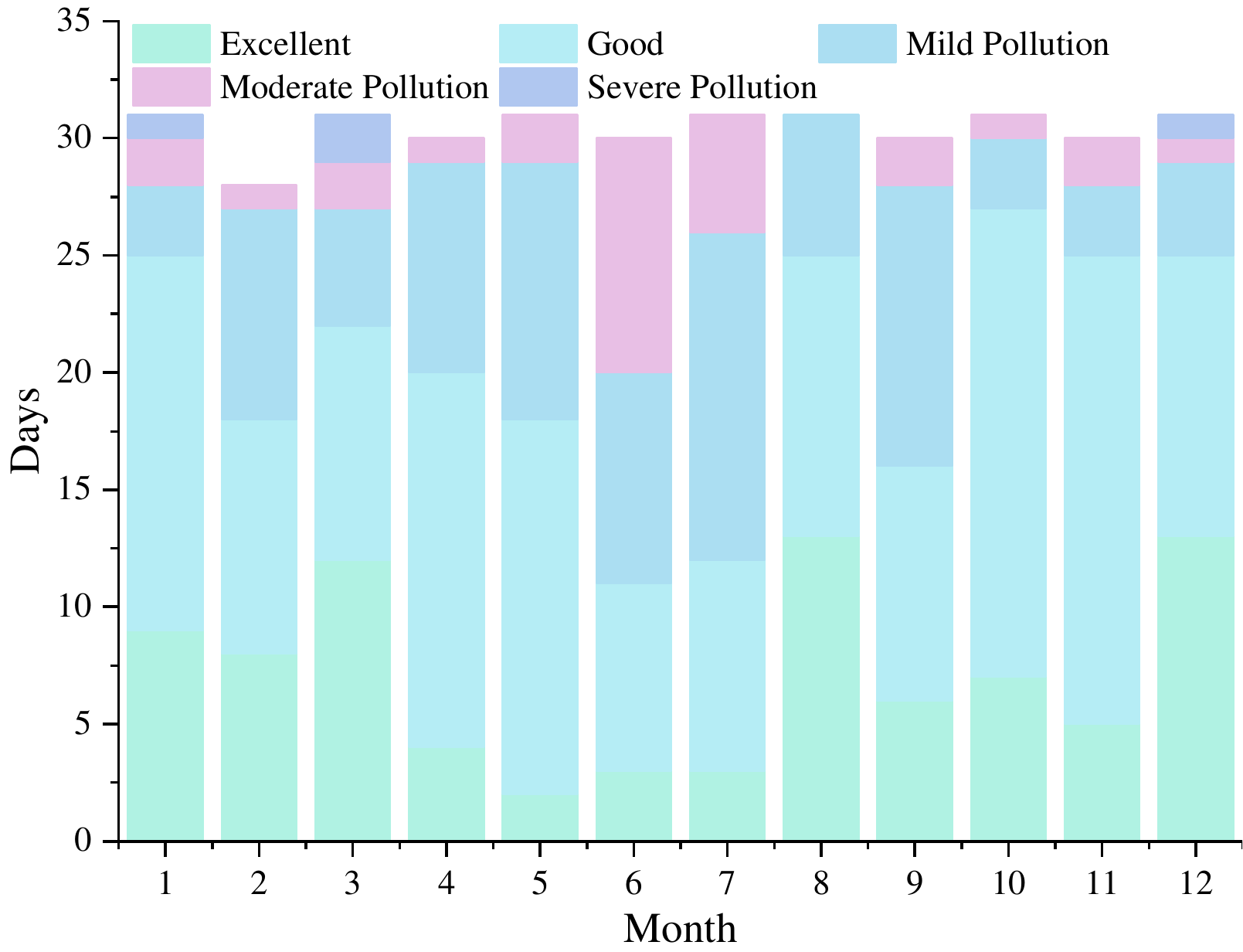}
        \label{AQIbeijing}
    }
    \\
    \vspace{1em}
    \subfigure[\textcolor{black}{Monthly variation of PM2.5 concentrations in a specific region of Jiangsu Province, 2015}]{
        \includegraphics[width=0.47\textwidth]{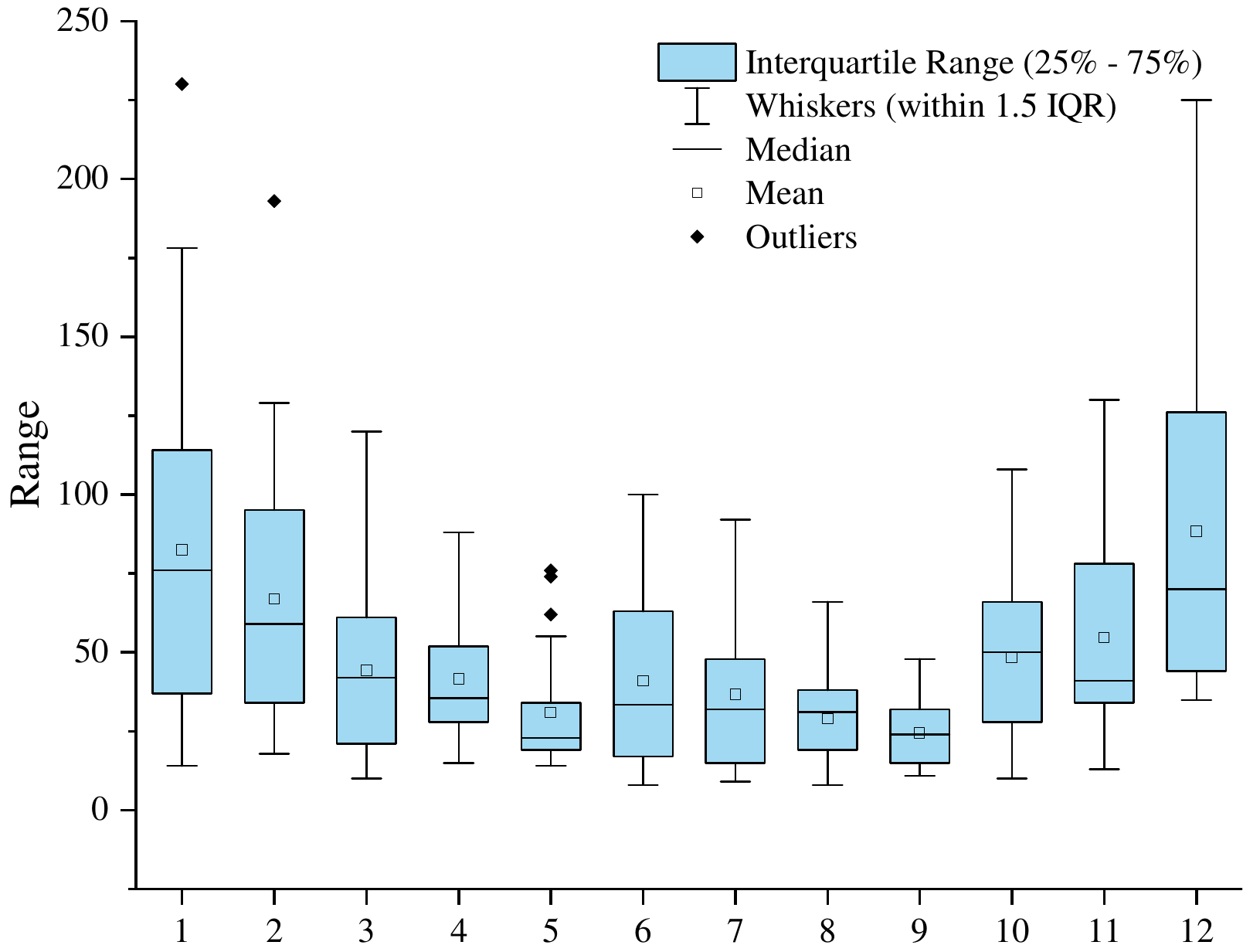}
        \label{PM2.5yan}
    }
    \subfigure[\textcolor{black}{Monthly variation of PM2.5 concentrations in Beijing, 2019}]{
        \includegraphics[width=0.47\textwidth]{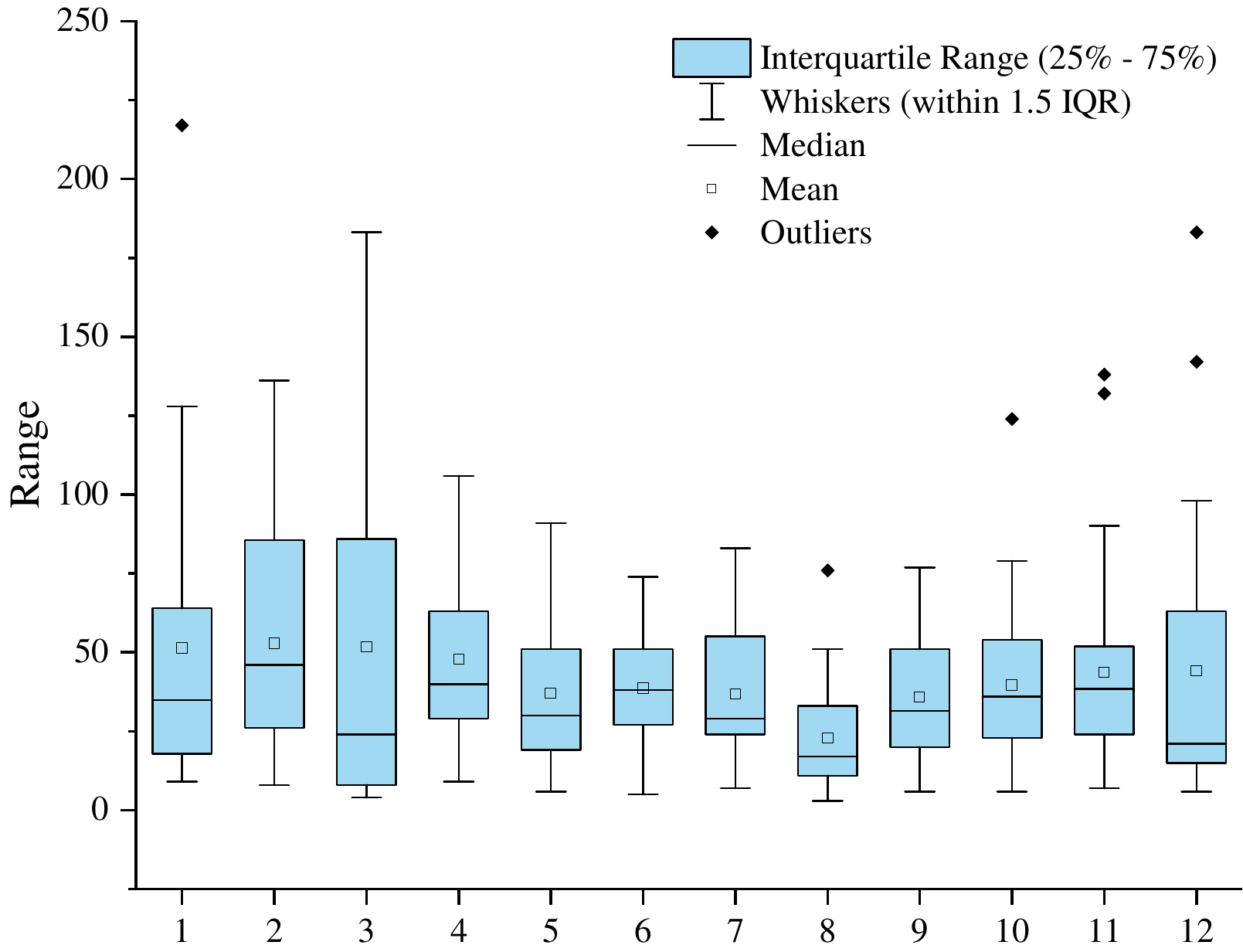}
        \label{PM2.5beijing}
    }
    \caption{\textcolor{black}{Analysis of Air Quality and PM2.5 Concentrations in Jiangsu Province (2015) and Beijing (2019)}}
    \label{fig:subfigures}
\end{figure*}

\subsection{Analysis of air quality}
\textcolor{black}{In this section, we analyze the air quality of a specific region in Jiangsu Province in 2015 and the air quality in Beijing in 2019. The air quality datasets for both locations can be found at https://www.aqistudy.cn/historydata/.}

\textcolor{black}{Fig. \ref{AQIyan} and \ref{AQIbeijing} show the monthly air quality days for a region in Jiangsu and Beijing, respectively. It can be observed that there are more days with severe pollution in January and December, while the air quality is better in the summer. Fig. \ref{PM2.5yan} and \ref{PM2.5beijing} present the monthly variation of PM2.5 concentrations in the same regions. PM2.5 concentrations fluctuate significantly in winter while they remain relatively stable in summer.}

\textcolor{black}{From the four figures, we can conclude that air quality in both datasets exhibits significant seasonal variations, with more severe pollution levels typically occurring during the winter months, particularly in January. The data shows that PM2.5 concentrations in January are highly variable, with a wide range of values and numerous outliers, indicating unstable and often high pollution levels. }

\textcolor{black}{Based on the above analysis, we selected the January data from the two datasets for single-step forecasting experiments. For multi-step forecasting experiments, we used the Beijing data from January to March, where the first 12 points were used to predict the subsequent 4 points.}

\begin{figure*}[h]
    \centering
    \includegraphics[width=0.75\linewidth]{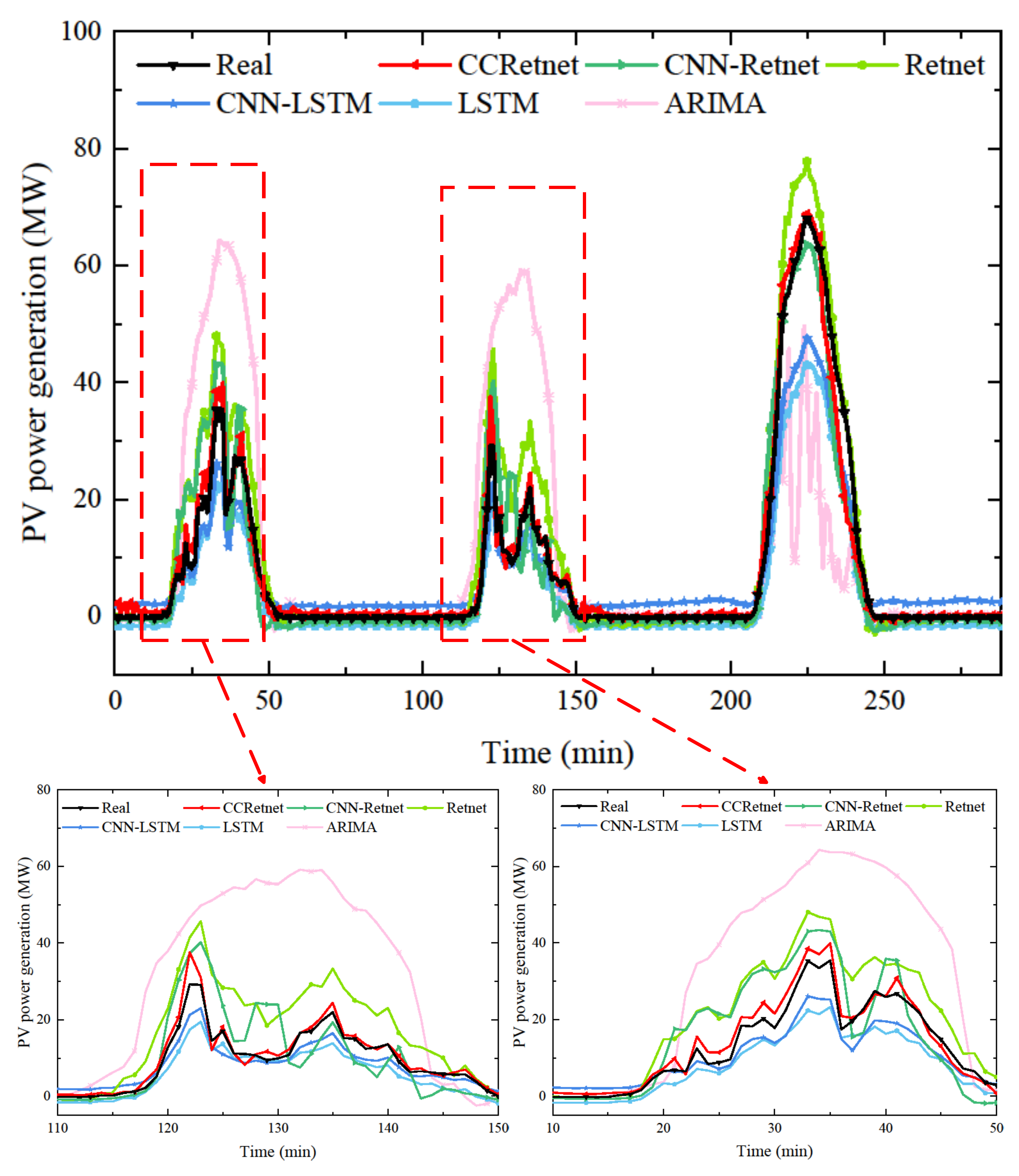}
    \caption{\textcolor{black}{Comparison of PV power generation forecasting on different models in a specific region of Jiangsu Province, January 2015}}
    \label{result1}
\end{figure*}
\begin{figure*}[H]
    \centering
    \subfigure[\textcolor{black}{Evaluation metrics of different models in a specific region of Jiangsu Province, January 2015}]{
        \includegraphics[width=0.45\textwidth]{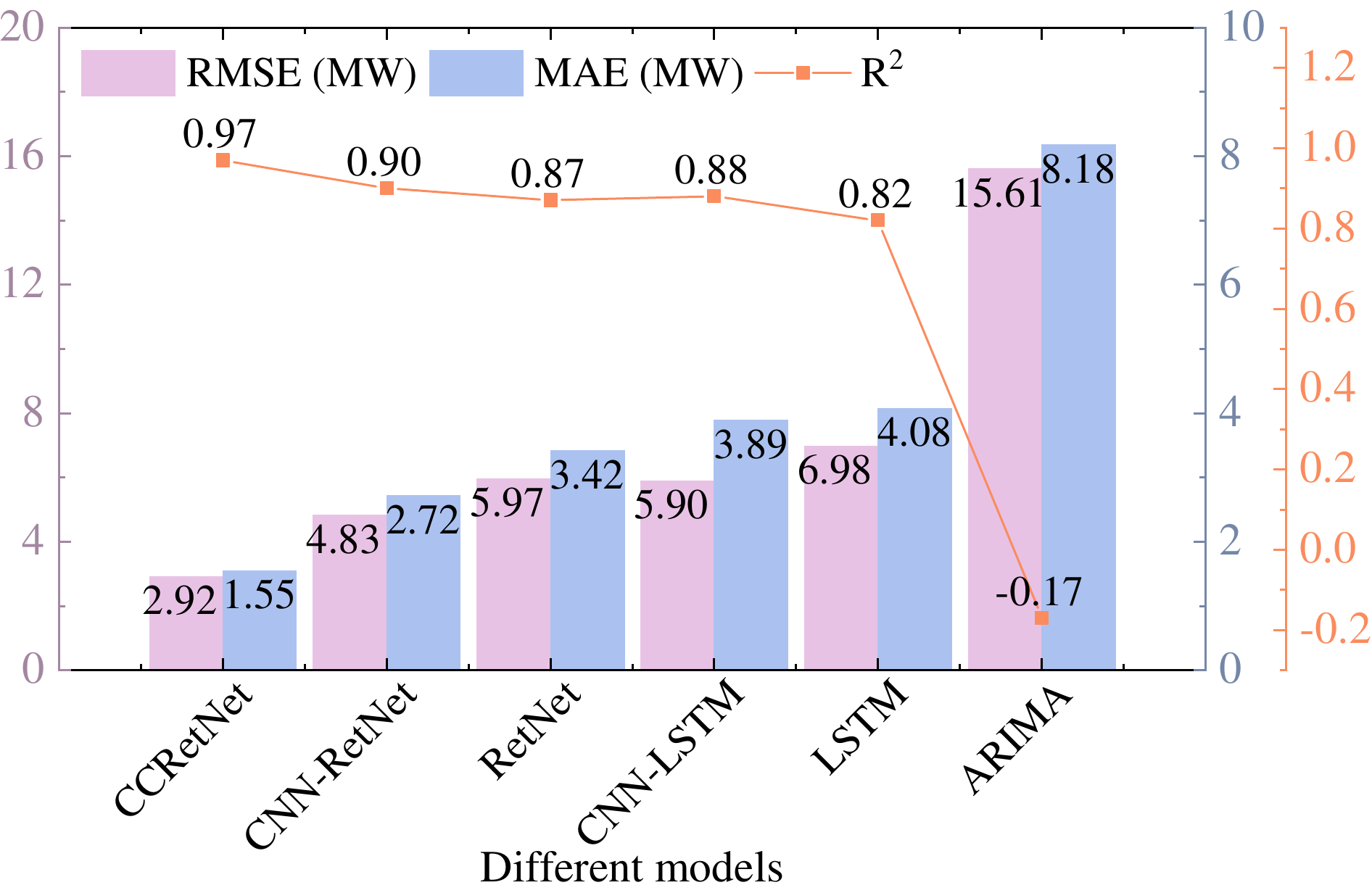}
        \label{error}
    }
    \subfigure[\textcolor{black}{Evaluation metrics of different optimization methods on CCRetNet in a specific region of Jiangsu Province, January 2015}]{
        \includegraphics[width=0.4\textwidth]{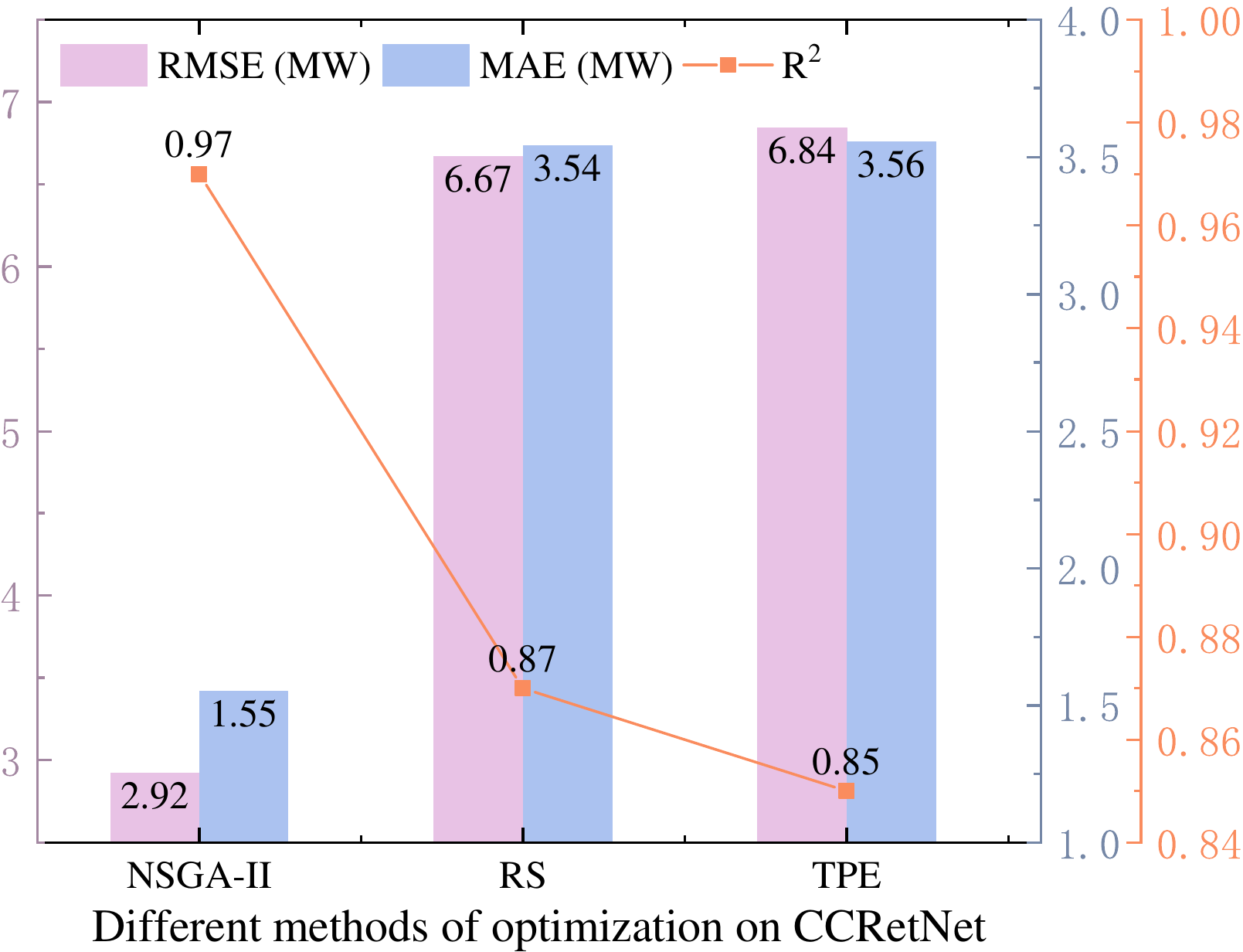}
        \label{optimize}
    }
    \caption{\textcolor{black}{Comparison of evaluation metrics and optimization methods for PV power generation forecasting in Jiangsu Province, January 2015}}
\end{figure*}

\twocolumn 
\clearpage 

\subsection{Experimental results and analysis}
\textcolor{black}{We conduct experiments using two datasets, with the results discussed in Section \ref{single-stepJiangsu Province}, Section \ref{single-stepBeijing}, and Section \ref{multi-stepBeijing}.}

\textcolor{black}{We compare the proposed model with eight traditional models, including CCRetNet, CNN-Retnet, Retnet, CNN-LSTM, CNN-GRU, CNN-RNN, long short term memory (LSTM), gate recurrent unit (GRU), recurrent neural network (RNN), autoregressive integrated moving average model (ARIMA) and Transformer. We utilize the Optuna library for hyperparameters and choose three methods for comparison: NSGA-II, random search (RS), and tree-structured parzen estimator (TPE). Model performance is evaluated based on accuracy and computational costs. }

\subsubsection{Analysis of single-step forecasting experiments in Jiangsu Province, 2015} \label{single-stepJiangsu Province}

\textcolor{black}{In this section, we emphasize the experimental outcomes of multiple forecasting models applied to a specific region of Jiangsu Province dataset.}

\begin{figure}
    \centering
    \includegraphics[width=0.7\linewidth]{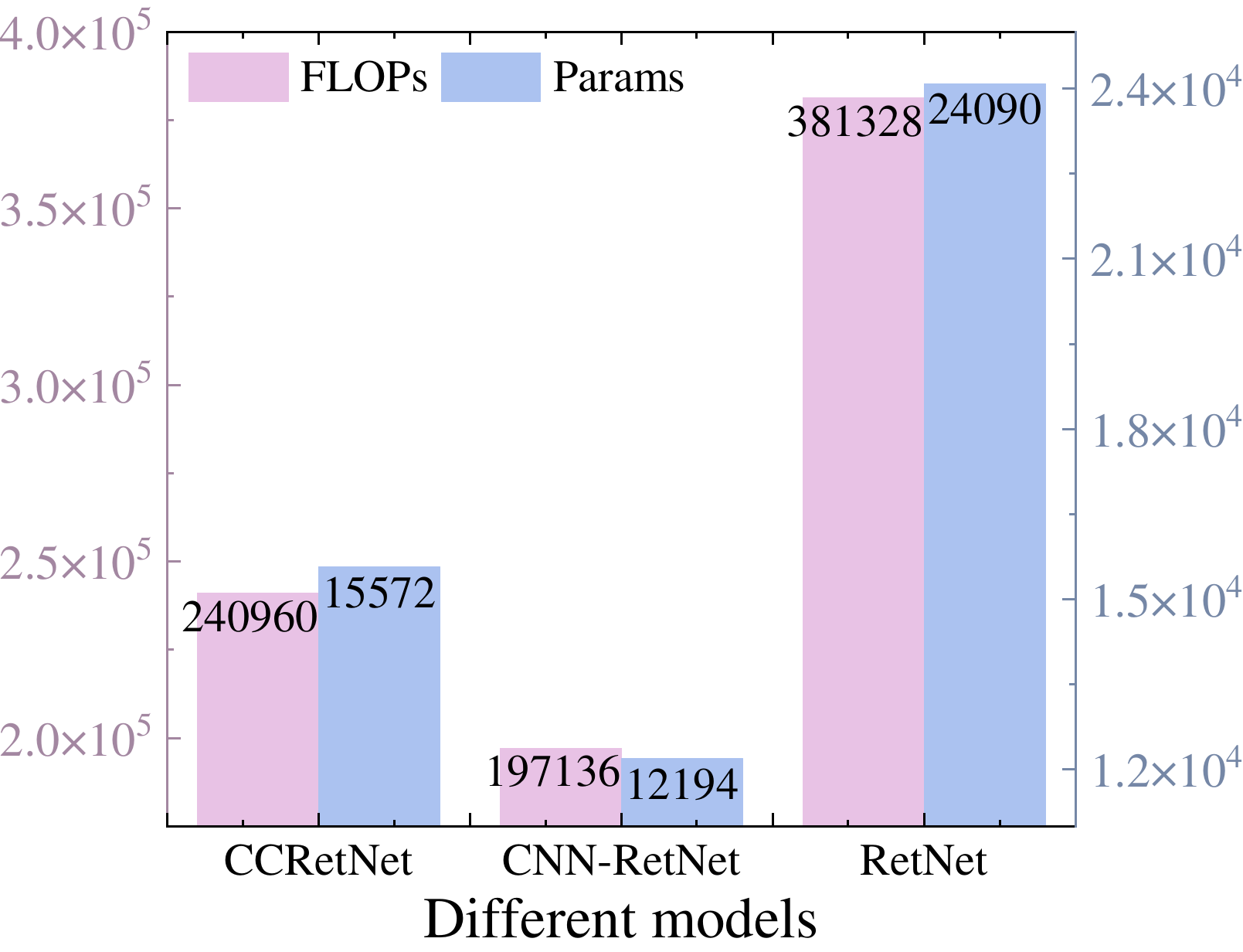}
    \caption{\textcolor{black}{The computational costs on different models in a specific region of Jiangsu Province, January 2015}}
    \label{cost}
\end{figure}

\textcolor{black}{We compare the forecasting performance of various models, highlighting the superiority of CCRetNet in handling significantly fluctuating real data. Fig. \ref{result1} presents the forecasting results, while Fig. \ref{error} provides numerical analysis. CCRetNet performs best when real data fluctuates considerably, with the lowest RMSE and MAE values of 2.71 MW and 2.55 MW and $R^2$ value of 0.97, higher than the other models, demonstrating its superiority. In contrast, ARIMA performs poorly when real data fluctuates significantly, with a negative $R^2$ value, indicating that it is unsuitable for PV forecasting on this dataset, especially during hazy weather.}

\textcolor{black}{The ablation experiments further verify the superiority of CCRetNet. Fig. \ref{result1} and Fig. \ref{error} show that CCRetNet outperforms CNN-RetNet, primarily due to the clustering method based on TEWPP, which effectively qualifies uncertainty during hazy weather and classifies them within the clustering module. Fig. \ref{result1} and Fig. \ref{error} also indicate that CNN-RetNet performs better than RetNet due to the introduction of CNN. The effectiveness of CNN is further supported by comparisons with other CNN-based deep learning models and general deep learning models.}

\textcolor{black}{Fig. \ref{optimize} shows the accuracy of CCRetNet under different optimization methods across three evaluation metrics. 
The model optimized by NSGA-II outperforms the others. 
The model optimized by RS also shows commendable performance with a robust $R^2$ value, suggesting a reliable predictive capability. 
The model optimized by TPE appears less effective, as it exhibits the highest RMSE and MAE of others.}

\textcolor{black}{Fig. \ref{cost} illustrates the computational costs of the three models. RetNet incurs the highest computational costs. Although CCRetNet's computational costs are slightly higher than that of CNN-RetNet, the introduction of hierarchical clustering significantly improves CCRetNet's forecasting accuracy and trend performance.}

\subsubsection{Analysis of single-step forecasting experiments in Beijing, 2019} \label{single-stepBeijing}
\begin{figure}
    \centering
    \includegraphics[width=0.7\linewidth]{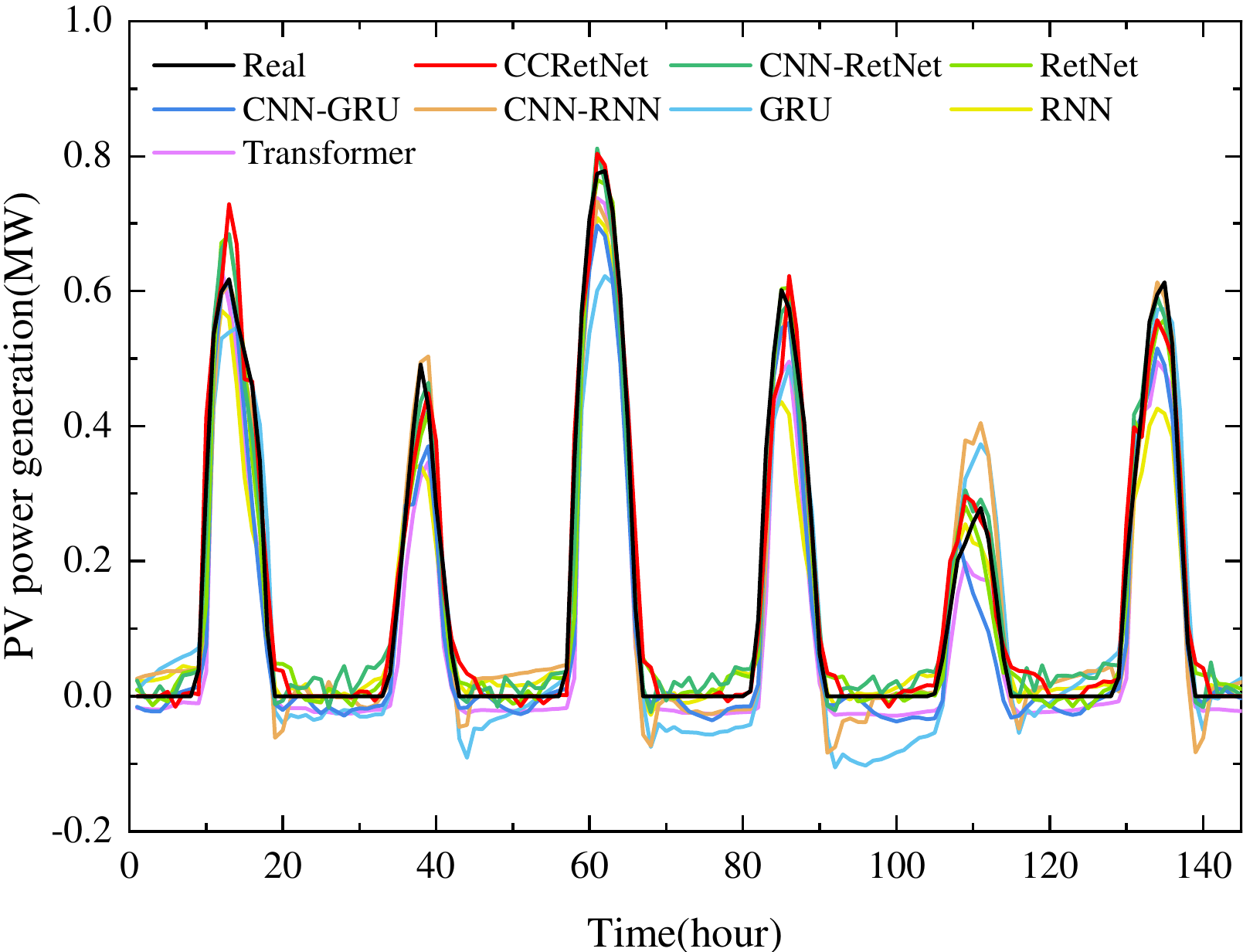}
    \caption{\textcolor{black}{Comparison of PV power generation forecasting on different models in Beijing, January 2019}}
    \label{real_beijingsingle}
\end{figure}

\textcolor{black}{This section presents the performance analysis of various forecasting models on the Beijing dataset, highlighting both trend alignment and numerical accuracy.}

\textcolor{black}{Fig. \ref{real_beijingsingle} compares the actual values and the forecasts of traditional models on the Beijing dataset for January 2019. Overall, the $R^2$ values indicate that traditional models perform better on the Beijing dataset than on the Jiangsu dataset. CCRetNet exhibits minimal deviation from actual results during stable periods, although its performance during peak values is comparable to other traditional models as shown in Fig. \ref{real_beijingsingle}.}

\begin{figure*}[H]
    \centering
    \subfigure[\textcolor{black}{Evaluation metrics of different models in Beijing, January 2019}]{
        \includegraphics[width=0.47\textwidth]{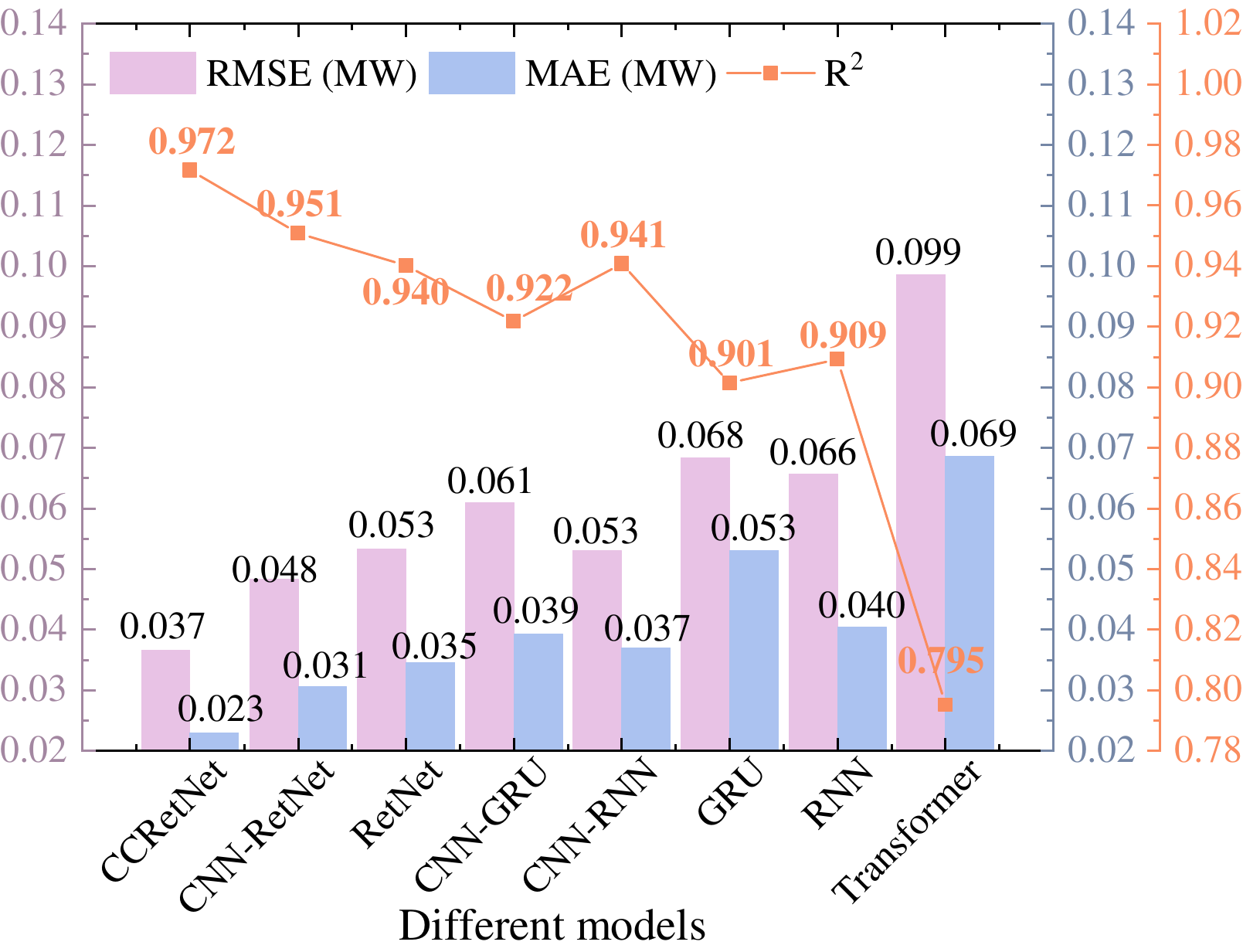}
        \label{error1_beijingsingle}
    }
    \subfigure[\textcolor{black}{Evaluation metrics of different models in Beijing, March 2019}]{
        \includegraphics[width=0.47\textwidth]{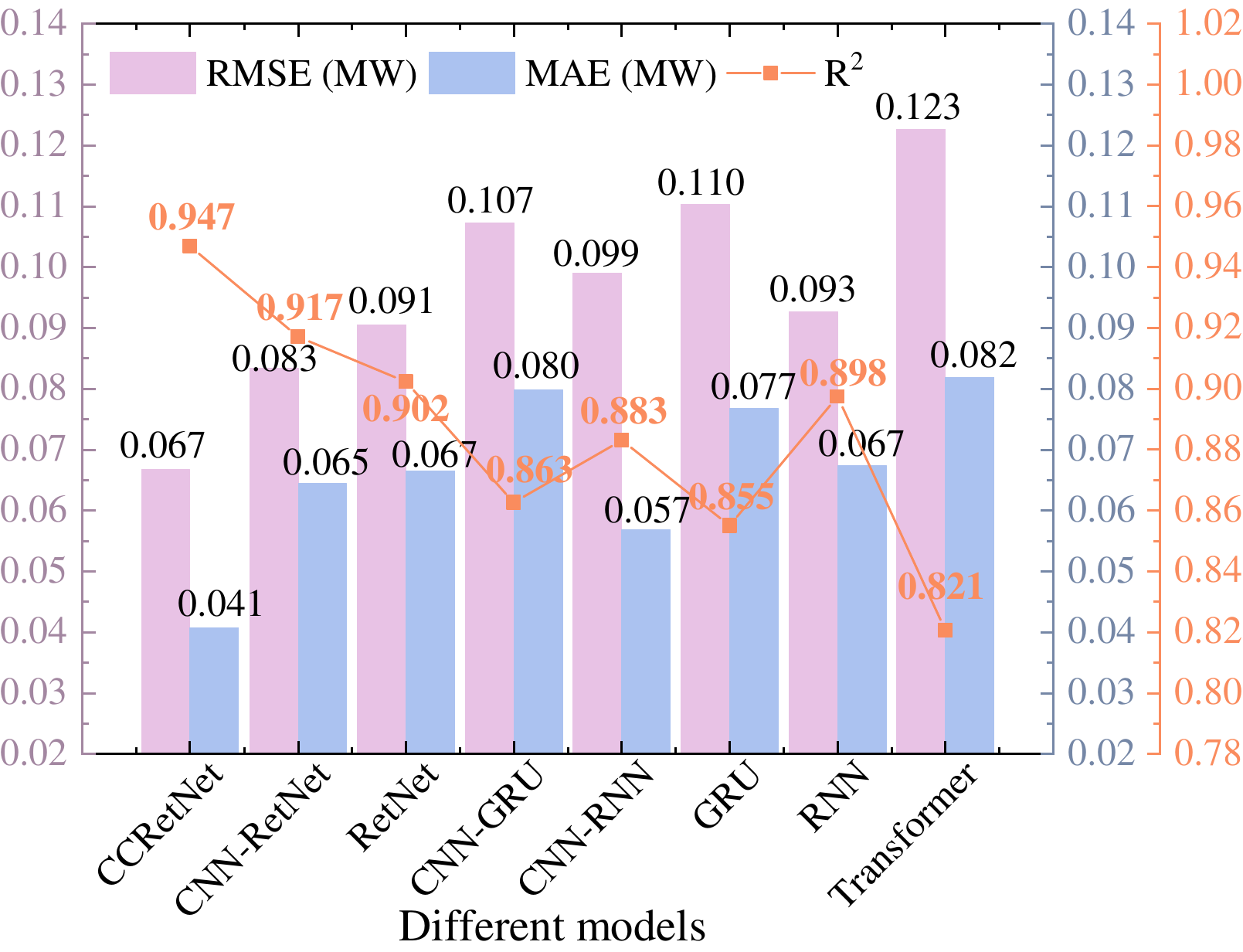}
        \label{error1_beijingmulti}
    }
    \\
    \subfigure[\textcolor{black}{Evaluation metrics of different optimization methods on CCRetNet in Beijing, January 2019}]{
        \includegraphics[width=0.47\textwidth]{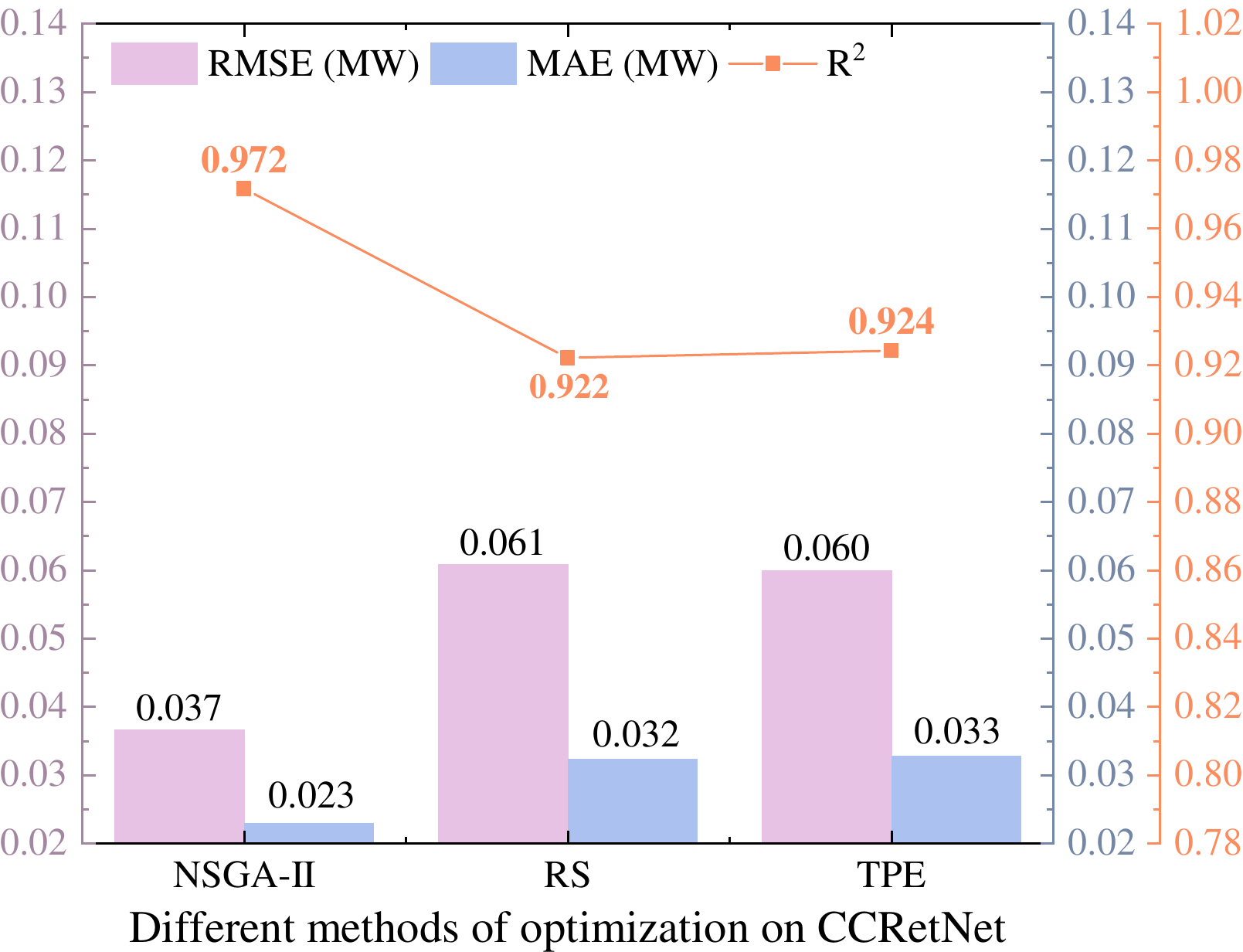}
        \label{optimize_beijingsingle}
    }
    \subfigure[\textcolor{black}{Evaluation metrics of different optimization methods on CCRetNet in Beijing, March 2019}]{
        \includegraphics[width=0.47\textwidth]{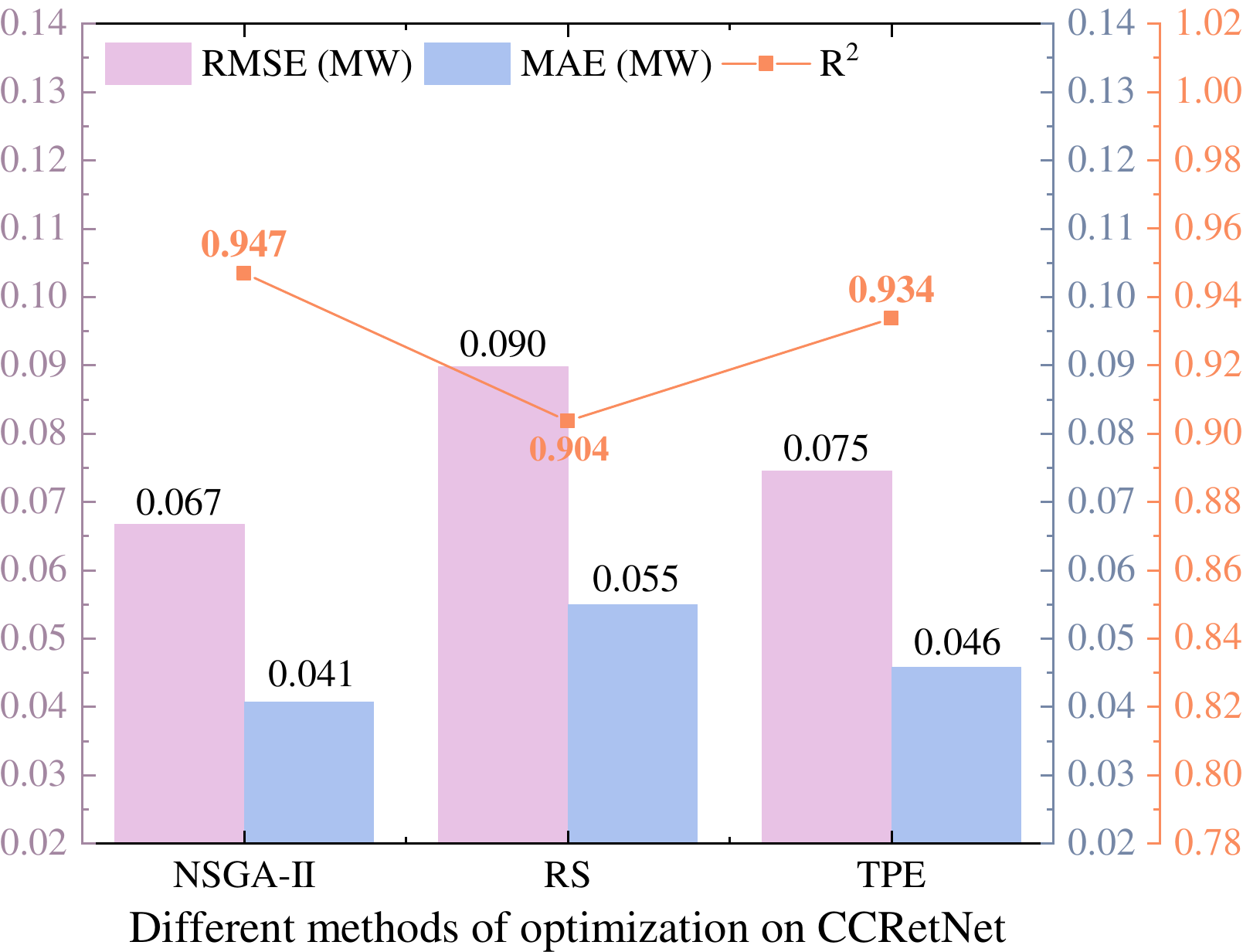}
        \label{optimize_beijingmulti}
    }
    \caption{Comparison of evaluation metrics and optimization methods for PV power generation forecasting in Beijing, January and March 2019}
\end{figure*}

\textcolor{black}{Moreover, Fig. \ref{error1_beijingsingle} reveals that CCRetNet achieves the smallest deviation from actual results, with the ablation experiment further confirming its superiority over CNN-RetNet, particularly due to the effectiveness of TEWPP-based clustering in qualifying uncertainty.}

\textcolor{black}{Fig. \ref{optimize_beijingsingle} compares the performance of CCRetNet under different optimization methods. The model optimized by NSGA-II achieves the best forecasting accuracy, evidenced by the lowest RMSE and MAE. The models optimized by TPE and RS both outperform the NSGA-II optimized model. However, they show similar forecasting accuracy on the Beijing dataset, which differs from the results on the Jiangsu dataset.}

\begin{figure*}
    \centering
    \includegraphics[width=0.95\linewidth]{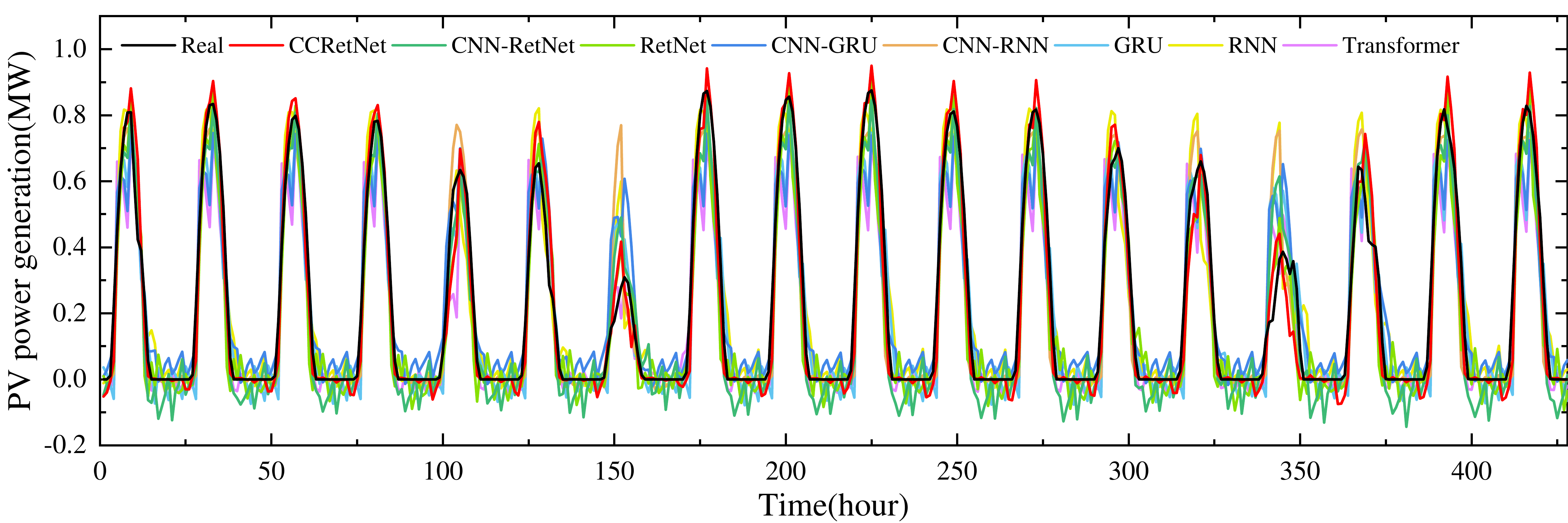}
    \caption{\textcolor{black}{Comparison of PV power generation forecasting on different models in Beijing, March 2019}}
    \label{real_beijingmulti}
\end{figure*}
\vspace{1em}
\twocolumn 
\clearpage 

\subsubsection{Analysis of multi-step forecasting experiments in Beijing, 2019}\label{multi-stepBeijing}

\textcolor{black}{In this section, we present the experimental results of several multi-step forecasting models on the Beijing dataset.}

\textcolor{black}{In multi-step forecasting experiments, most models perform worse than in single-step forecasting, as shown in Fig. \ref{real_beijingmulti}, Fig. \ref{error1_beijingmulti}, and Fig. \ref {optimize_beijingmulti}. 
Although the MSE and MAE of CNN-BiLSTM and Transformer are higher in multi-step forecasting, the $R^2$ is closer to 1 compared to single-step forecasting.}

\textcolor{black}{In the ablation experiments, CCRetNet exhibits smaller forecasting fluctuations compared to other models when PV power generation is stable, similar to the results of single-step forecasting. The ablation experiments also demonstrate the superiority of the clustering method based on TEWPP and CNN method.}

\textcolor{black}{The CCRetNet model optimized by NSGA-II performs the best in multi-step forecasting, comparing optimization methods. As shown in Fig. \ref{optimize_beijingmulti}, the model optimized by TPE shows better forecasting accuracy than the model optimized by RS. However, both are outperformed by the model optimized by NSGA-II, a result that contrasts with the previous two experiments.}

\section{Conclusion} \label{conclusion}
\textcolor{black}{This study focuses on qualifying uncertainty in a series of PV power generation during hazy weather. 
The effectiveness of the proposed model is validated with two datasets.
We create TEWPP to qualify uncertainty during hazy weather.
Additionally, we use a novel median linkage method for hierarchical clustering to reduce computational costs. }
The optimal number of clusters $k$ is determined using the silhouette coefficient. 
Multiple CCRetNet models are trained and tested for various categories and hyperparameters optimized by the NSGA-II algorithm.
\textcolor{black}{The results indicate that CCRetNet outperforms other comparative models regarding RMSE, MAE, and $R^{2}$ for short-term forecasting.}

\textcolor{black}{The study focuses on PV forecasting during hazy weather without considering other weather scenarios.} Therefore, our future research aims to create a more efficient and adaptable framework by focusing on year-round weather conditions to make forecasts better.



\printcredits

\bibliographystyle{model1b-num-names}
\bibliography{cas-refs2}

\vspace{10cm}
\section*{Appendix}
An example of weather data is as follows:

\begin{figure}[h]
    \centering
    \includegraphics[width=1\linewidth]{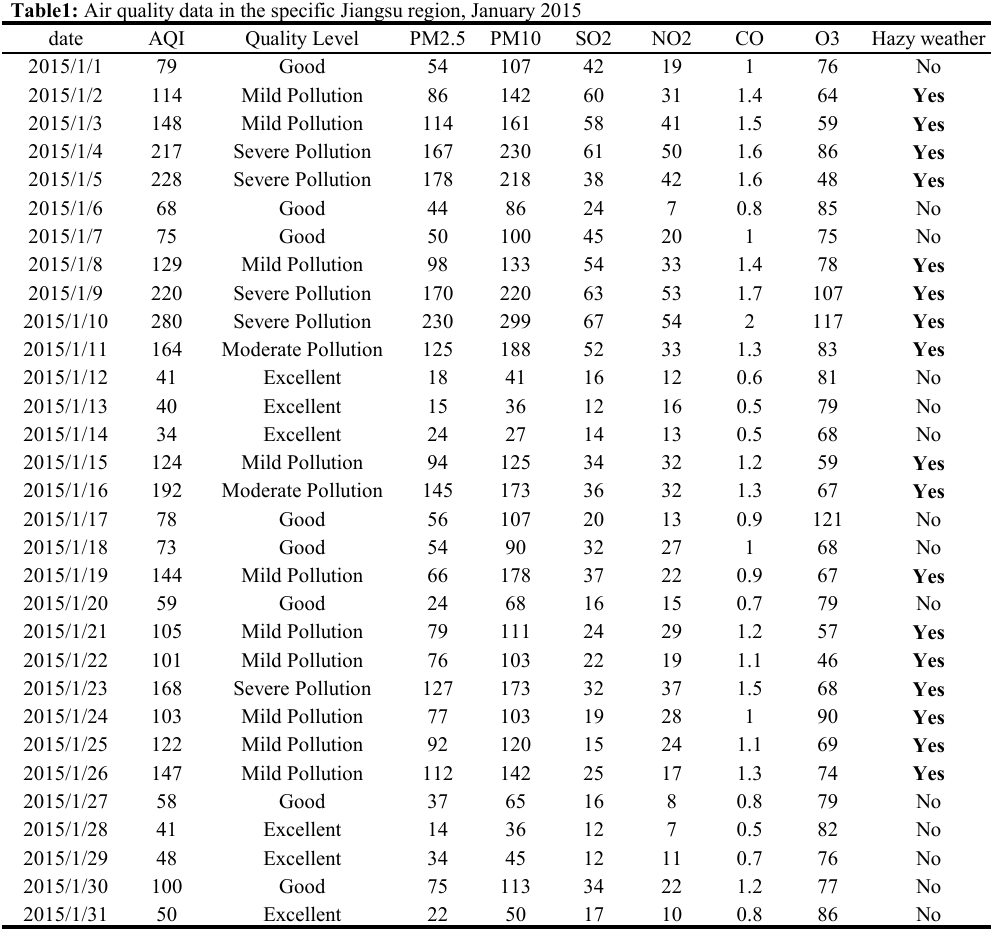}
    \label{fig:enter-label}
\end{figure}

\end{document}